\documentclass[11pt]{elsarticle} 
\usepackage{amssymb}
\usepackage{graphicx}%
\usepackage{multirow}%
\usepackage{amsmath,amssymb,amsfonts}%
\usepackage{amsthm}%
\usepackage{mathrsfs}%
\usepackage[title]{appendix}%
\usepackage{xcolor}%
\usepackage{textcomp}%
\usepackage{manyfoot}%
\usepackage{booktabs}%
\usepackage{listings}%
\usepackage{hyperref}

\usepackage{threeparttable}
\usepackage{tabularx}
\usepackage{bbm}
\usepackage{float}

\usepackage{graphicx}
\usepackage{subfigure}
\usepackage{natbib}

\usepackage{mathtools} % to get ceil{}

\DeclarePairedDelimiter\floor{\lfloor}{\rfloor}

\newcommand{\Ntest}{N_{\text{test}}}

\makeatletter
\def\ps@pprintTitle{%
  \let\@oddhead\@empty
  \let\@evenhead\@empty
  \let\@oddfoot\@empty
  \let\@evenfoot\@oddfoot
}
\makeatother

\begin{document}

\begin{frontmatter}

%% Title, authors and addresses

%% use the tnoteref command within \title for footnotes;
%% use the tnotetext command for theassociated footnote;
%% use the fnref command within \author or \affiliation for footnotes;
%% use the fntext command for theassociated footnote;
%% use the corref command within \author for corresponding author footnotes;
%% use the cortext command for theassociated footnote;
%% use the ead command for the email address,
%% and the form \ead[url] for the home page:
%% \title{Title\tnoteref{label1}}
%% \tnotetext[label1]{}
%% \author{Name\corref{cor1}\fnref{label2}}
%% \ead{email address}
%% \ead[url]{home page}
%% \fntext[label2]{}
%% \cortext[cor1]{}
%% \affiliation{organization={},
%%            addressline={}, 
%%            city={},
%%            postcode={}, 
%%            state={},
%%            country={}}
%% \fntext[label3]{}

%% use optional labels to link authors explicitly to addresses:
%% \author[label1,label2]{}
%% \affiliation[label1]{organization={},
%%             addressline={},
%%             city={},
%%             postcode={},
%%             state={},
%%             country={}}
%%
%% \affiliation[label2]{organization={},
%%             addressline={},
%%             city={},
%%             postcode={},
%%             state={},
%%             country={}}

\title{Uncertainty quantification in automated valuation models with spatially weighted conformal prediction}

\author[1,2]{Anders Hjort\corref{cor1}}
\ead{anderdh@math.uio.no}

\author[1]{Gudmund Horn Hermansen}
\ead{gudmunhh@math.uio.no}

\author[1]{Johan Pensar}
\ead{johanpen@math.uio.no}

\author[3]{Jonathan P. Williams}
\ead{jwilli27@ncsu.edu}

\affiliation[1]{
organization = {Department of Mathematics, University of Oslo}, 
addressline = {0851 Oslo}, 
country = {Norway}
}

\affiliation[2]{
organization = Eiendomsverdi AS, 
addressline = {0104 Oslo}, 
country = {Norway}
}

\affiliation[3]{
organization = {Department of Statistics, North Carolina State University}, 
addressline = {27695 Raleigh}, 
state = {North Carolina}, 
country = {USA}
}

\cortext[cor1]{Corresponding author. \\ Visiting Address: Moltke Moes vei 35, Niels Henrik Abels hus, 0851 Oslo.\\ Postal Address: Postboks 1053 Blindern, 0316 Oslo. \\ Email: anderdh@math.uio.no. \\ Telephone: +47 995 82 778.}

%%%%%%%%%%%%%%%%%%%%%%%%%%%%%%%%%%%%%%%%%%%%%%%%%%%%%%%%%%%%%%%%%%%%%%%%%%%%%%

\begin{abstract}
Non-parametric machine learning models, such as random forests and gradient boosted trees, are frequently used to estimate house prices due to their predictive accuracy, but a main drawback of such methods is their limited ability to quantify prediction uncertainty. Conformal prediction (CP) is a model-agnostic framework for constructing confidence sets around predictions of machine learning models with minimal assumptions. However, due to the spatial dependencies observed in house prices, direct application of CP leads to confidence sets that are not calibrated everywhere, i.e., the confidence sets will be too large in certain geographical regions and too small in others. We survey various approaches to adjust the CP confidence set to account for this and demonstrate their performance on a data set from the housing market in Oslo, Norway. Our findings indicate that calibrating the confidence sets on a spatially weighted version of the non-conformity scores makes the coverage more consistently calibrated across geographical regions. We also perform a simulation study on synthetically generated sale prices to empirically explore the performance of CP on housing market data under idealized conditions with known data-generating mechanisms.

\end{abstract}


\end{frontmatter}

\section{Introduction}
\label{sec:introduction}

The housing market plays a crucial role in most modern economies, and buying a home is the most significant financial investment many people make throughout their lives. Homeowners, real estate professionals, and financial institutions, such as banks and insurance companies, therefore need to monitor the value of a home regularly. While manual appraisal often results in an accurate valuation, performing it regularly on a portfolio of thousands of homes is infeasible. Therefore, participants in the real estate market rely on Automated Valuation Models (AVMs), typically in the form of statistical models that estimate the value of a home given its location, characteristics, and previous sale price. Modern AVMs often rely on non-parametric machine learning approaches like gradient boosted trees, random forests, or a combination of both \cite{SteurerHillPfeifer2021}. These models have been shown to deliver superior predictive accuracy compared with more classical approaches, such as linear regression, due to their flexible and non-parametric nature while also being able to handle a combination of numerical and categorical features, which are often present in housing market data sets.

Despite their predictive accuracy, the models mentioned above often lack inherent mechanisms for quantifying the uncertainty in their predictions. Practitioners are often interested in not only a point prediction but a range of plausible values, referred to as a confidence set. Returning a confidence set in addition to the point prediction makes it easier to evaluate the trustworthiness of the AVM and use it for risk-management purposes. Empirical data indicates that house prices display significant heteroscedasticity and spatial heterogeneity \cite{Marques2021} and that even the state-of-the-art AVMs yield very different accuracy across different spatial regions and price levels \cite{Hjort2022}, further motivating the need for calibrated uncertainty quantification in AVMs. 

Conformal prediction (CP) is a popular framework for uncertainty quantification in the field of machine learning and pattern recognition. The underlying method of CP was introduced in \cite{ALRW} as a distribution-free test of exchangeability, but it has recently gained popularity within the pattern recognition community as a general-purpose method for constructing confidence sets around a prediction \citep[see][and the references therein]{Toccaceli2022}. The recent popularity has seen CP being used in a wide range of applications, including in active learning \cite{Matiz2019}, rule extraction \cite{Johansson2022}, and multi-target regression \cite{Messoudi2021}.

The fundamental idea behind the CP framework is to assess whether a test data point is exchangeable with a set of training observations, relying on a non-conformity score that quantifies how dissimilar or non-conforming each data point is to the training set. From the training set, we can use these scores as a reference to calibrate our expectations about the uncertainty of a new, unobserved instance. Specifically, we create a confidence set for the test instance by only including values that conform sufficiently with the training set at some user-specified confidence level. An appealing trait of the CP framework lies in its generality. The confidence sets created by the CP procedure are distribution-free, with few assumptions on the point prediction used in the regression. Furthermore, the confidence sets are \textit{marginally valid} in the following sense: A $90\%$ confidence set created by a CP algorithm on a training set has a $90\%$ probability of covering a test instance under the assumption that the test instance is exchangeable with the training data. We thus expect $90\%$ of exchangeable test observations to be covered by the confidence sets created by the CP framework without knowing the distribution of the data.  

%To make the CP methods more useful and trustworthy in real-life ML applications, it is sometimes desirable with 

In real-life applications to housing data, it is often desirable to construct CP sets that are not only marginally valid but also provide the correct coverage in different subsets of the feature space, referred to as approximate conditional coverage. In recent years, several methodological adjustments have been made to the CP framework in an attempt to achieve approximate conditional coverage. A notable class of methods to achieve this is weighted versions of CP, where the CP sets for a new test point are calibrated against a weighted version of the non-conformity scores. Instead of treating all the non-conformity scores as equally important when calibrating the confidence sets, higher weight is assigned to non-conformity scores collected close to the test instance, where closeness can be defined either in geographical space \citep{MartinMaoReich}, feature space \citep{Guan}, or time \citep{BeyondExchangeability}.  Investigating methods for approximate conditional coverage in strata of the spatial domain is particularly appealing to the real estate application, as the underlying spatial processes that drives the prices is complex, and it is often challenging to correctly specify a point prediction model or a non-conformity score to account for this. As such, this research aims to investigate the use of spatially weighted CP to create CP sets that are not only marginally valid but also yield approximately conditional coverage, i.e., the correct coverage within the geographical strata present in the data. 

We are not the first to use CP for real estate applications. In \cite{Bellotti2017}, CP is applied to a data set from the London housing market, focusing on adjusting for temporal drifts in house prices. Furthermore, specific non-conformity scores for AVMs are developed in \cite{Bellotti} to construct CP sets that are as narrow as possible. The pursuit of approximate conditional coverage across geographical regions is also the topic in a study of the Norwegian housing market in \cite{Hjort2024}. However, their approach fundamentally differs from ours, as they identify clusters of municipalities based solely on the empirical distribution of non-conformity scores without incorporating the spatial dimension. A CP method based on quantile regression is also applied to an AVM setting in \cite{BastosPaquette2024} on a data set from San Francisco, USA, where the authors also study the approximate conditional coverage across different strata of the feature space.  

While these studies provide important insights into the construction of CP sets for AVMs, we are, to the best of our knowledge, the first to demonstrate the effectiveness of spatially weighted non-conformity scores in AVMs. We show how the weighting concept can be incorporated with different choices of non-conformity scores and point prediction models to improve the performance of the CP sets. To evaluate the approach, we study a novel data set of $N = 26\,362$ transactions from the housing market in Oslo, Norway. The case study demonstrates that assigning a higher weight to non-conformity scores in a spatial neighborhood yields confidence sets that adapt better to local spatial patterns compared to standard unweighted CP.

The rest of the paper is structured as follows. \autoref{sec:avms} introduces the application of AVMs, \autoref{sec:oslo} introduces the data set from Oslo that will be studied, \autoref{sec:cp} presents the necessary theoretical framework on CP and its extensions based on weighted non-conformity scores. \autoref{sec:results} presents the results of the considered methods in two scenarios: First with synthetically generated sale prices and then on the actual transaction prices observed in the Oslo data set. Concluding remarks are made in \autoref{sec:conclusion}.

\section{Automated valuation models}
\label{sec:avms}

Buying a home is often among the most significant financial decisions people make throughout their lives. Most home buyers need to take out a mortgage to finance the acquisition, and The Organisation for Economic Co-operation and Development (OECD) reports that the average Loan-to-Income rate in OECD countries was approximately $4.4$ in 2018 \citep{OECD_MortgageFinance}. It is, therefore, necessary for both banks and homeowners to regularly monitor the value of a home or a portfolio of homes. While manual appraisal through physical inspections is one way of estimating the home value, these are often time-consuming and hard to conduct regularly. For these reasons, many financial institutions rely on AVMs to provide statistical estimates of housing values. 

%While multiple approaches to constructing AVMs exist, the most common models are \textit{hedonic pricing models} that estimate the price of a dwelling given its characteristics, such as the location, size, number of bedrooms, etc. Early examples of regression models being used for this purpose can be seen in \cite{bailey1963regression} and \cite{rosen1974hedonic}, and recent years have seen an increased use of non-parametric machine learning models like random forests (\cite{BreimanRF}) and gradient boosted trees (\cite{XGBoost}) as hedonic price models in AVMs (\cite{SteurerHillPfeifer2021}).

While multiple approaches to constructing AVMs exist, the most common models are \textit{hedonic pricing models} that estimate the price of a dwelling given its characteristics, such as the location, size, number of bedrooms, etc. Recent years have seen an increased use of non-parametric machine learning models like random forests \citep{BreimanRF} and gradient boosted trees \citep{XGBoost} as hedonic price models in AVMs \citep{SteurerHillPfeifer2021}. The popularity of non-parametric tree-based models can be attributed to, amongst other things, their ability to detect higher-order interactions and their ability to work with both categorical and numerical data. A decision tree recursively partitions the feature space into a set of non-overlapping regions $\mathcal{R}_1, ..., \mathcal{R}_L$, where each region is represented by a leaf or terminal node in the tree structure. Each region, $\mathcal{R}_l$, takes a value that serves as the prediction for any test instance whose feature values place it in $\mathcal{R}_l$. Although a single decision tree might be limited in its predictive power, ensemble methods like random forests or gradient boosted trees often achieve superior predictive accuracy in many prediction tasks on tabular data. %Certain tree-based regression methods can be interpreted as a form of nearest neighbor regression (\cite{LinJeon2006}). 

AVMs typically suffer from considerable omitted-variable bias, with vital information such as the floor plan, the degree of interior luxury, or the dwelling's exterior design typically being unknown or hard to quantify. There are examples from the pattern recognition literature of image recognition software being used to recognise interior of dwellings \citep{Shallow2Deep2020}, but this is rarely reliably available for an entire portfolio of dwellings. Other sources of influence on the final sale price might include weather patterns around the day of the open house or bidding strategies in the auction leading up to the sale, which both might be challenging to account for a priori. %As a consequence of the inherent uncertainty and noise in the prediction task, AVMs are often evaluated using the Percentage Error Range (PER), which measures the fraction of predictions on a test set that is within either $\pm10\%$ or $\pm20\%$ of the actual sale price \cite{SteurerHillPfeifer2021}. This industry-specific performance metric indicates that even predictions that under- or overestimate by more than $10\%$ may be deemed acceptable due to the challenging nature of the prediction task. 

The increased deployment of machine learning methods in AVMs calls for increased attention to uncertainty quantification in AVMs. The non-parametric nature of the models mentioned above comes with the consequence that they are rarely equipped with straightforward methods for creating calibrated prediction intervals. Although several industry white papers exist on evaluating confidence sets for AVMs, there is a lack of universally agreed-upon metrics to quantify and assess the different uncertainty quantification methods \citep{Krause2020}. Academic research on specific methods for creating prediction intervals in AVMs is also scarce, although some methods exist. An application to house price prediction with quantile regression is presented in \cite{Hao2023}, whereas \cite{DeepForest2020} use a classification algorithm on a set of pre-defined price bins. In the realm of conformal prediction (CP), \cite{Bellotti2017} provide a case study from the housing market in London, with a particular focus on handling temporal changes in housing prices when constructing CP sets. In \cite{Bellotti}, the authors develop a set of novel non-conformity scores based on normalized absolute residuals. The authors point out that the absolute residuals of AVMs tend to correlate positively with the price the AVM tries to predict. In \cite{BastosPaquette2024}, a CP method based on quantile regression is employed on a data set from San Francisco, US, and the conditional coverage properties are studied. 

%A dichotomization between \textit{model-based} and \textit{error-based} uncertainty models for AVMs is introduced in \cite{Krause2020}. Model-based uncertainty measurements are derived directly from the underlying point prediction, often relying on parametric models (like linear regression) with some inherent uncertainty quantification. In contrast, the error-based uncertainty models use historical errors to assign confidence to new predictions. ## \textcolor{red}{Better in thesis}

\section{The Oslo data set}
\label{sec:oslo}
We study a data set of transactions from the housing market in Oslo, Norway's capital and largest city. The data set consists of $N = 26\,362$ transactions, each representing a single sale between a buyer and a seller. The data set is provided by Eiendomsverdi AS, a financial technology company from Norway. All transactions are from the years $2016-2017$ and only include transactions of apartments, as opposed to detached or semi-detached homes. The data set includes the sale price, measured in million Norwegian kroner (NOK)\footnote{1 NOK $\approx$ 0.09 USD per September 2024.}, which is the transaction price agreed upon by a seller and buyer after negotiations, not the initial ask price. The data set also includes $p = 17$ dwelling-specific covariates. Examples of such covariates are size, number of bedrooms, and floor. There are two variables encoding information about the size of the dwellings: size and gross size. The first encodes the livable area, which includes bedrooms, bathrooms, kitchens, and living rooms. The latter, gross size, refers to the usable area, which includes the livable area and additional areas such as basements or storage rooms. Both are measured in square meters. 

The location of the dwelling is encoded via the coordinates (longitude and latitude) but also by a city district variable. There are $15$ city districts in Oslo. In addition, the ``Lake Distance'' and ``Coast Distance'' variables encode information about the relative location. Similarly, the variables ``Homes Nearby'' and ``Other Buildings Nearby'' provide information about the number of residential and commercial buildings close to each dwelling, based on a discretization of Norway into grids of $250\times 250$ meter. The temporal trends in the data set is encoded through the ``Sale Month'' variable, which ranges from $1$ to $24$, since we study a two year period. A data summary is provided in \autoref{table:covariates}. 

\begin{table}
\centering 

\begin{threeparttable}
\footnotesize
\begin{tabular}{@{}l c c c c c c@{}}
%\hline \\[-1.8ex] 
\toprule
Variable & \multicolumn{1}{c}{Unit}  & \multicolumn{1}{c}{Mean} & \multicolumn{1}{c}{St. Dev.} & \multicolumn{1}{c}{Min} & \multicolumn{1}{c}{Max} & \multicolumn{1}{c}{Type}  \\ 
%\hline \\[-1.8ex] 
\midrule
Sale Price & NOK (mill.) &  4.28 & 1.84 & 0.35 & 27.9 & Numerical \\
Size  & $m^2$ & 65.68 & 24.43 & 12 & 343 & Numerical\\ 
Gross Size & $m^2$ & 67.20 & 25.29 & 12 & 368  & Numerical \\ 
Longitude & degrees & 10.78 & 0.06 & 10.62 & 10.95 & Numerical \\ 
Latitude & degrees & 59.92 & 0.03 & 59.82 & 59.98 & Numerical\\ 
City District\tnote{a} & - & - & - & - & - & Categorical \\
Altitude & $m$ & 94.93 & 62.21 & 0 & 480  & Numerical \\ 
Bedrooms & - & 1.78 & 0.76 & 0 & 8 & Numerical\\ 
Floor\tnote{b} & - & 3.03 & 2.07 & $-$4 & 99& Numerical\\ 
Age & years &  58.55 & 36.71 & 0 & 266  & Numerical \\ 
Coast Distance & $m$ & 3,261 & 2,453 & 5 & 12,201 & Numerical\\ 
Lake Distance & $m$ &  993 & 504 & 26 & 3,018 & Numerical\\ 
Balcony\tnote{c} & - & 0.75 & 0.43 & 0 & 1 & Categorical \\ 
Elevator\tnote{c} & - & 0.36 & 0.48 & 0 & 1 & Categorical \\ 
Sale Month & - & 12.25 & 6.87 & 1.0 & 24.0 & Categorical \\ 
Units On Address\tnote{d} & - & 20.86 & 29.12 & 0 & 274 & Numerical\\
Homes Nearby\tnote{e}& - & 2,721 & 1,586 & 98 & 6,746 & Numerical\\ 
Other Buildings Nearby\tnote{e}& - & 153.0 & 135.9 & 6.0 & 1243.0 & Numerical\\ 
\bottomrule
\end{tabular} 
\begin{tablenotes}
  \footnotesize
  \item[a] There are $15$ distinct city districts in the data set.
  \item[b] If the dwelling has multiple floors, this variable will be the lowest floor. 
  \item[c] In cases where the information is missing, this is set to $0$. 
  \item[d] In apartment buildings, multiple dwellings often have the same address.
  \item[e] Norway is divided into squares of $250 \times 250$ meters. This variable counts the number of homes or other buildings (stores, schools, churches) in all the adjacent squares to the square where the targeted dwelling is, i.e., the $8$ neighboring squares.
\end{tablenotes}

\caption{The variables in the data set with summary statistics for the numerical variables.} 
\label{table:covariates} 
\end{threeparttable}

\end{table}

\begin{figure}
    \centering
    \includegraphics[width = 0.99\linewidth]{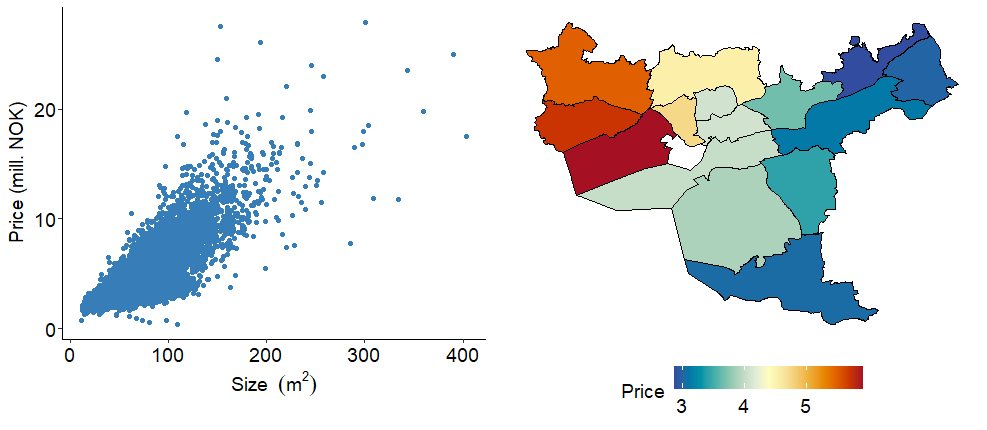}
    \caption{Visualizations of the Oslo $2016-2017$ data set. \textbf{Left:} Sale price plotted against the size in $m^2$. \textbf{Right:} Mean price per in million NOK per city district. The white city district has no observations.} 
    \label{fig:descriptive_data}
\end{figure}

\autoref{fig:descriptive_data} displays how the sale price varies with the size of the dwelling (left panel) and how the average price varies per city district in Oslo (right panel). These two attributes, the size and the location, are often among the most important characteristics for a home buyer, and typically serve as significant predictors for the final sale price. Not surprisingly, the mean price increases with the size of the dwelling. It should be noted that while homes up to $400$ square meters are visualized in \autoref{fig:descriptive_data}, more than $90\%$ of the dwellings are $100$ square meters or less, and the median size is $62$ square meters.

Significant variability is observed across regions, as the northwestern city districts display prices that are, on average, above NOK $5$ million. In comparison, the city districts located further east have mean prices around NOK $3$ million. The city district colored in white in \autoref{fig:descriptive_data} is Sentrum, a city district without observations in the data set. The absence of transactions in this area reflects that office buildings and tourist attractions dominate this city district with little residential housing.

\section{Conformal prediction}
\label{sec:cp}
CP is a distribution-free and model-agnostic framework for the construction of confidence sets at a pre-specified confidence level $\alpha \in (0,1)$ for unobserved test data, given a sequence of already observed training data from the same albeit unknown distribution. This section introduces how CP can be used to create confidence sets in regression settings and presents a literature survey on ways to adjust the CP framework when the test instance is believed to come from a different distribution than the training data. 

\subsection{Simple illustration}
\label{subsec:cp}
Assume that we have observed a sequence of random variables $Z_{1:N} := (Z_1, ..., Z_N)$ with  $Z_i \in \mathbb{R}$ for every $i$, and seek to create a confidence set for an unobserved $Z_{N+1} \in \mathbb{R}$ given that it comes from the same distribution as $Z_{1:N}$. The CP approach \citep{ALRW} quantifies uncertainty by hypothesizing a value $Z_{N+1} = z$ and then calculating a conformal p-value for this hypothesis. To calculate a p-value without making distributional assumptions about the data-generating process, CP utilizes a \textit{non-conformity} measure that quantifies how dissimilar the hypothesized value $z$ is from the previously observed $Z_{1:N}$. Consider, for instance, the non-conformity measure
\begin{align}
    \Psi_i(Z_i) := \lvert Z_i - \Bar{Z}_{-i}\rvert, 
    \label{eq:hold_out_score}
\end{align}
where $\Bar{Z}_{-i}$ is the mean of $\{Z_{1:(N+1)} \}\setminus \{Z_i\}$. A high value of the non-conformity measure for $Z_{N+1}$ indicates that $Z_{N+1}$ is dissimilar to the training data, while a low value indicates a high degree of similarity. We then calculate a non-conformity score for each data point in the sequence $(Z_1, ..., Z_{N+1}=z)$. We denote the realized non-conformity scores $s_1, ..., s_{N+1}$, with $s_i := \Psi_i(Z_i)$ as in \eqref{eq:hold_out_score}. Under the null hypothesis that the hypothesized value $z$ is compatible with $Z_{1:N}$, then the rank of its non-conformity score $s_{N+1}$ is uniformly distributed among $s_1, ..., s_n$. A p-value for the hypothesis that $z$ comes from the same distribution as $Z_{1:N}$ can then be constructed as 
\begin{align*}
    p_{N+1} = \frac{1}{N+1}\sum_{i = 1}^{N+1} \mathbbm{1}\{s_{N+1} \leq s_{i}\}.
\end{align*}

We accept the hypothesis if $p_{N+1} > \alpha$ for some confidence level $\alpha \in (0,1)$. Define $\hat{q}_{1-\alpha}$ to be the $(1-\alpha)$th percentile of $s_1,..., s_N$, i.e.,  
\begin{align*}
    P(s_{N+1} \leq \hat{q}_{1-\alpha}) \geq 1-\alpha
\end{align*}
under the null hypothesis. We can now invert the conformal p-value over all possible values $z$ and construct a CP set $C_{1-\alpha} \subset \mathbb{R}$ that contains all the hypothesized $z$ values that are accepted at level $\alpha$ based on the conformal p-value. Formally, the CP set is thus given by
\begin{align*}
    C_{1-\alpha} &= \big\{ z: \Psi_{N+1}(z) \leq \hat{q}_{1-\alpha} \big\}.
 \end{align*}

\subsection{Inductive conformal prediction}

The procedure described above, where the mean $\Bar{Z}_{-i}$  is recalculated for every $i$ through a hold-out approach, is referred to as Transductive CP \citep{ALRW} or Full CP \citep{lei2017distributionfree}. This method is often computationally infeasible since the mean must be recomputed $N+1$ times for every hypothesized $Z_{N+1} = z$. A method to circumvent the recalculation of the mean (in the case of our simple illustration) for every score, referred to as Split CP, consists of splitting the full data set into two parts: A training set of size $m$ and a calibration set of size $n$, such that $m+n = N$. In this setting, the training set can be used to calculate the mean $\Bar{Z}_m$ once, and the non-conformity measure is rewritten as
\begin{align*}
    \Psi_i(Z_i) := \lvert Z_i - \Bar{Z}_m\rvert,
\end{align*}
with the training mean $\Bar{Z}_m$ replacing the hold-out mean $\Bar{Z}_{-i}$ used in the Transductive CP non-conformity score from \eqref{eq:hold_out_score}. The reduction in computational complexity comes at the cost of a reduction in statistical efficiency, however, potentially resulting in larger CP sets \citep{lei2017distributionfree}, since the number of non-conformity scores used in the calibration step is reduced. The mean estimate $\Bar{Z}_m$ is is also less efficient since $m < N$. Since smaller CP sets are typically preferred, the Transductive CP approach is generally favorable, but the Inductive CP provides a more practical and computationally quicker alternative that is reasonable in situations where the data set is large enough to set aside a calibration set of sufficient size.

\subsection{Extension to regression} 
We now expand to a regression context where the goal is to create a CP set for an unobserved $Y_{N+1} \in \mathbb{R}$ given a feature vector $X_{N+1} \in \mathbb{R}^d$. For notational simplicity we denote $Z_i := (X_i, Y_i) \in \mathbb{R}^d \times \mathbb{R}$, and only present the Split CP method. 

As in Split CP, we assume the existence of a training set $(Z_1, ..., Z_{m})$ and a calibration set $(Z_{m+1}, ..., Z_N)$. The training set is used to train a prediction model $\hat{f}: \mathbb{R}^d \to \mathbb{R}$, with no particular restrictions on the class of models or the algorithm used to train $\hat{f}$ other than that it is invariant to the order of the training data \citep{lei2017distributionfree}. The canonical non-conformity measure in this context is $\Psi_i: \mathbb{R}^d \times \mathbb{R} \to \mathbb{R}$, with 
\begin{align*}
    \Psi_i(Z_i) := \lvert Y_i - \hat{f}(X_i)\rvert
\end{align*}
being a widely used choice \citep{ALRW}. Although $X_i$ is now of $d$ dimensions, the non-conformity score is still defined in $\mathbb{R}$.

When a model is trained, and a non-conformity score is defined, we then use these to score all the observations in the calibration set and obtain realized non-conformity scores $s_{1}, ..., s_{n}$. When given an instance $X_{N+1} \in \mathbb{R}^d$ and a pre-specified confidence level $\alpha \in (0,1)$, the CP set $C_{1-\alpha}(X_{N+1})$ is defined as 
\begin{align}
    C_{1-\alpha}(X_{N+1}) &= \big\{ y: \Psi_{N+1}(X_{N+1}, y) \leq \hat{q}_{1-\alpha} \big\}
 \label{eq:cp_set}
\end{align}
where  $\hat{q}_{1-\alpha}$ is the $(1-\alpha)$th empirical quantile of $s_1, ..., s_n$. 

In practice, inverting the conformal p-value for every value of $y$ is a computational bottleneck. It is thus demonstrated in \cite{lei2017distributionfree} that when the non-conformity measure is the absolute residual (as described here), and we use the Split CP algorithm, the CP set can be expressed as
\begin{align}
    C_{1-\alpha}(X_{N+1}) = [\hat{f}(X_{N+1}) \pm \hat{q}_{1-\alpha}], 
    \label{eq:cp_set2}
\end{align}
reducing the computational burden.\footnote{To see how this reduces, observe that $P(s_{N+1} \leq \hat{q}_{1-\alpha}) \geq 1 - \alpha$ under the null hypothesis that $Z_{N+1}$ is exchangeable with $(Z_1, ..., Z_N)$. In the Split CP setting, $s_{N+1}$ does not rely on a recomputation of $\hat{f}$, so this is equivalent to writing $P( \lvert Y_{N+1} - \hat{f}(X_{N+1})\rvert) \leq \hat{q}_{1-\alpha}) \geq 1 - \alpha$, which equates to $P(\hat{f}(X_{N+1}) - \hat{q}_{1-\alpha} \leq Y_{N+1} \leq \hat{f}(X_{N+1}) + \hat{q}_{1-\alpha}) \geq 1 - \alpha$.}

\subsection{Coverage guarantees}
If we create $C_{1-\alpha}(X_{N+1})$ as described in \eqref{eq:cp_set}, the CP set comes with a \textit{marginal coverage guarantee}: 
\begin{align}
    \mathbb{P}\big\{Y_{N+1} \in C_{1-\alpha}(X_{N+1})\big\} \geq 1-\alpha. 
    \label{eq:marginal_guarantee}
\end{align}
CP sets that yield this guarantee are referred to as being marginally valid \citep{ALRW}. The guarantee in \eqref{eq:marginal_guarantee} holds regardless of the underlying distribution as long as the data is exchangeable. A sequence $(Z_1, ..., Z_{N+1})$ is exchangeable if its joint probability density is invariant to permutations on the order. 

Although the marginal coverage guarantee is useful, the guarantee is averaged across a whole test set.\footnote{The coverage guarantee is also valid across the calibration set. However, it is often mostly of interest to study the behavior of the CP sets on new instances.} A more desirable guarantee is \textit{conditional coverage}: 

\begin{align*}
    \mathbb{P}\big\{Y_{N+1} \in C_{1-\alpha}(X_{N+1})\big| X_{N+1} = x \big\} \geq 1-\alpha, 
\end{align*}
where the coverage holds conditionally for every possible value $X_{N+1} = x$. From a practical point of view, a conditional coverage guarantee is typically more useful than a marginal coverage guarantee, but it is demonstrated in \cite{LeiWasserman2014} that conditional coverage is only possible if the prediction sets $C_{1-\alpha}(X_{N+1})$ are infinitely large. Since exact conditional coverage guarantees are infeasible, a milder requirement is \textit{approximate conditional coverage}, 
\begin{align}
    \mathbb{P}\big\{Y_{N+1} \in C_{1-\alpha}(X_{N+1})\big| X_{N+1} \in \mathcal{X}_k \big\} \geq 1-\alpha, \quad \text{for} \quad k = 1, ..., K,
    \label{eq:approximate_conditional_guarantee}
\end{align}
where $\mathcal{X}_1, ..., \mathcal{X}_K$ is a partition of the feature space. The partition might be an inherent feature of the data set, such as geographical regions, or it can also be defined in more sophisticated ways by binning feature values or clustering observations. Mondrian CP is proposed in \cite{ALRW} as a straightforward way of achieving approximate conditional coverage. The Mondrian CP framework calculates the empirical quantile of interest separately for each subset $\mathcal{X}_k$. Variations of the approximate conditional coverage guarantee in  \eqref{eq:approximate_conditional_guarantee} are presented in various ways in the literature with slight differences in how the subsets $\mathcal{X}_k$ are constructed. We refer to \cite{LimitsOfDistributionFreeCoverage} for a detailed discussion of \eqref{eq:approximate_conditional_guarantee}, including lower bounds on the size of the subsets $\mathcal{X}_k$.

%%%%%%%%
%%%%%%%%
%%%%%%%%
\subsection{The role of the non-conformity measure}
\label{subsec:ncm}
The choice of non-conformity measure plays an important role in the construction of the CP sets. If $Z_{1:N}$ are exchangeable, as required by the CP procedure, any choice of non-conformity measure leads to marginally valid CP sets \citep{ShaferVovk2007}. However, a well-constructed non-conformity measure can lead to more \textit{efficient} sets, meaning sets that are smaller on average. Given the choice between two different procedures for CP set construction that have the same theoretical coverage guarantees, smaller set sizes are preferred. 

The non-conformity measure proposed so far leads to CP sets of constant size in the Split CP regression setting. In the regression context of \eqref{eq:cp_set2}, \cite{Papadopoulos2002} propose an adaptive non-conformity measure:  
\begin{align*}
    \Psi (Z_i) = \frac{\lvert Y_i - \hat{f}(X_i)\rvert}{\hat{\sigma}(X_i)},
\end{align*}
where $\hat{\sigma}$ serves as a normalizing function. With this non-conformity measure, the CP sets take the form  
\begin{align*}
    C_{1-\alpha}(X_{N+1}) = [\hat{f}(X_{N+1}) \pm \hat{\sigma}(X_{N+1})\cdot \hat{q}_{1-\alpha}],  
\end{align*}
i.e., it is adjusted based on the value of $X_{N+1}$ \citep{lei2017distributionfree, Papadopoulos2002}. As opposed to \eqref{eq:cp_set2}, which creates CP sets of constant width, the $\hat{\sigma}$ function makes CP sets of adaptive size. The function serves as an estimate of the Mean Absolute Deviance of the quantity $Y-\hat{f}(X)\lvert X = x$ \citep{lei2017distributionfree}, and is referred to as the \textit{difficulty} function in \cite{Bostrom2017}. Estimation of the difficulty function may rely on the inherent properties of the considered point predictor or characteristics of the data set and application at hand. Alternatively, the function can also be learned directly by fitting a model from the unnormalized non-conformity scores from the training set \citep{Papadopoulos2002}. 

Instead of using the absolute residual from the point prediction to build the non-conformity score, \cite{romano2019conformalized} propose a score based on quantile regression:
 \begin{align*}
    \Psi_i(Z_i) := \max \big\{ \hat{Q}_{\alpha/2} (X_i) - Y_i, Y_i - \hat{Q}_{1-\alpha/2}(X_i) \big\}, 
\end{align*}
where $\hat{Q}_{\alpha/2}$ and $\hat{Q}_{1-\alpha/2}$ are the estimates of the $\alpha/2$:th and $(1-\alpha/2)$:th quantile from a quantile regression. The CP sets for a test instance are then constructed as $C_{1-\alpha}(X_{N+1}) = [\hat{Q}_{\alpha/2}(X_{N+1}) - \hat{q}_{1-\alpha}, \hat{Q}_{1-\alpha/2}(X_{N+1}) + \hat{q}_{1-\alpha}]$, where $\hat{q}_{1-\alpha}$, as usual, is an empirical percentile of the non-conformity scores. This method is referred to as conformalized quantile regression (CQR). Unlike the other non-conformity scores presented, $\hat{q}_{1-\alpha}$ might take either positive or negative values, resulting in an expansion or shrinking of the conditional quantiles returned from the quantile regression method.

%%%%%%%%
%%%%%%%%
%%%%%%%%

\subsection{Spatially Weighted CP}
Recent years have seen several extensions of CP where the calculation of $\hat{q}_{1-\alpha}$ is conducted on a weighted sample of the non-conformity score, with the weights being a function of the test data point. This leads to some added computational complexity, as $\hat{q}_{1-\alpha}$ must be recomputed for every test instance, but results in CP sets that are potentially more adaptive to the test instance. 

One example of this is the Spatial CP framework presented in \cite{MartinMaoReich}. The authors propose to use a spatially weighted version of the non-conformity score: 
\begin{align*}
    w_{i} = \exp( - d_{i,N+1}^2/\eta),
\end{align*}
where  $d_{i, N+1}$ is the Euclidean distance between non-conformity score number $i$ and a test instance, and $\eta$ is a hyperparameter. The quantile of interest is thus a weighted quantile of the non-conformity scores $s_1, \dots, s_N$ with corresponding weights $w_1,\dots, w_N$. The authors also present a non-smooth version of the same idea with 
\begin{align*}
    w_{i} = \begin{cases}
        1 \quad \text{if} \quad d_{i, N+1} \leq D, \\ 
        0 \quad \text{else},
    \end{cases}
\end{align*}
for some threshold $D$. The authors demonstrate that both weighting methods yield marginal validity similar to unweighted CP methods with some assumptions on the data-generating mechanism. Under the same data-generating mechanisms, both methods also come with exact conditional coverage under an asymptotic infill regime. 

In \cite{Guan}, another framework based on weighted non-conformity scores is presented. This approach uses a localizer function that reweights non-conformity scores based on closeness in feature space. A challenge with this framework arises if the number of observations around the test instance is small, which is solved by adjusting the confidence level $\alpha$ individually for each test instance. This framework yields marginal coverage guarantees with finite samples and enjoys asymptotic conditional coverage guarantees in low dimensions. 

Weighted non-conformity scores have also been used to achieve marginal coverage guarantees even if the exchangeability assumption is relaxed. This has been explored for scenarios where the violation of the exchangeability is known \citep{CovariateShift} or unknown \citep{BeyondExchangeability}. Spatially weighted CP is applied in the latter, although not with the goal of obtaining stricter coverage guarantees, but rather to retain the marginal coverage of the CP method even when the non-conformity scores are not exchangeable. Finally, it is worth noting that Mondrian CP also fits into the weighting framework by assigning a weight of one to all the calibration data points from the same subset $\mathcal{X}_k$ as the test instance and a weight of zero otherwise. This also leads to subset-conditional coverage guarantees \citep{Vovk2012}.

\subsection{Evaluation metrics}
When evaluating CP sets on a test set $\mathcal{D}_{\text{test}} = (Z_{N+1}, \dots, Z_{N+\Ntest})$, we evaluate the coverage, i.e., the number of times a response from the test set is covered by the CP set. Specifically, the coverage gap between the theoretical and empirical coverage is defined to be  
\begin{align*}
    \text{Coverage-Gap}(\mathcal{D}_{\text{test}};\alpha) :=  (1-\alpha) - \frac{1}{\Ntest}\sum_{i = N+1}^{\Ntest} \mathbbm{1}\Big\{Y_i \in C_{1-\alpha}(X_i)\Big\}, 
\end{align*}
such that a positive coverage gap indicates too low empirical coverage, and a negative coverage gap indicates too high empirical coverage. We will expect to see greater variability in the empirical coverage when either the number of calibration instances and/or the number of test instances is low. In an extreme scenario with only one test instance, the empirical coverage is either $1$ or $0$. Theoretically, it is known \citep{Vovk2012} that the coverage follows a beta-binomial distribution that depend on the the number of calibration data points, the number of test data points, and the confidence level $\alpha$. For practical prediction problems this can be used as a diagnostic tool when evaluating the coverage gap. 

Furthermore, when studying the coverage properties in different strata of the data set, such as geographical regions, we consider the Mean Absolute Coverage Gap (MACG) across the subsets. Specifically, if the feature space (or geographical space) is divided into $K$ non-overlapping subsets, we evaluate 
\begin{align*}
    \text{MACG}(\mathcal{D}_{\text{test}}; 1-\alpha) = \frac{1}{K}\sum_{k = 1}^{K} \lvert  \text{Coverage-Gap}(\mathcal{D}_{\text{test}}^k; 1-\alpha) \rvert, 
\end{align*} 
where $\mathcal{D}_{\text{test}}^k$ is the test points in subset $k$. 

In addition to coverage, practitioners are often interested in studying the size of the CP sets, sometimes referred to as the \textit{inefficiency} of the sets, as larger sets are more inefficient than smaller sets. The efficiency of a CP set can either be measured in absolute terms or as a percentage of the true response. For an interval of the form $C_{1-\alpha}(X_i) = [L_i, U_i]$, define 
\begin{align*}
    \text{Inefficiency}\{C_{1-\alpha}(X_i), Y_i\} = (U_i - L_i)/Y_i,  
\end{align*}
i.e., the size of the set as a fraction of the true response $Y_i$. 

%The mean efficiency of the prediction intervals is
%\begin{align}
 %   \text{Eff}(X_1, ... ,X_{\Ntest}) = \frac{1}{\Ntest}\sum_{i = 1}^{\Ntest} \lvert C_{1-\alpha}(X_{i})\rvert 
 %   \label{eq:efficiency}
%\end{align}
%where $\lvert C_{1-\alpha}(X_{i})\rvert$ is the size of the prediction interval. We are also interested in the \textit{relative efficiency}, 
%\begin{align}
%    \text{rEff}(X_1, ... ,X_{\Ntest}) = \frac{1}{\Ntest}\sum_{i = 1}^{\Ntest} \frac{\lvert C_{1-\alpha}(X_{i})\rvert }{Y_{i}}, 
%    \label{eq:relative_efficiency}
%\end{align}
%where $Y_i$ is the true response. We measure the efficiency in the Norwegian currency, kroner. Narrower prediction intervals are usually preferred if the coverage is still $(1-\alpha)$. \
\section{Results}
\label{sec:results}

This section reports the results of different non-conformity scores and weighting methods on the Oslo data set presented in \autoref{sec:oslo}. Before considering the actual sale prices in the Oslo data set in \autoref{subsec:results_real_data}, we perform a simulation study in \autoref{sec:simulation_study} where the sale prices are synthetically generated from a known data-generating mechanism. The main focus of the simulation study is to investigate both the marginal and conditional coverage gap of the methods in three different scenarios on the distribution of noise in the data-generating model.  %The purpose of this is to compare and validate the methods in a controlled environment where the prediction uncertainty, which we seek to quantify via the CP framework, is explicitly controlled. 

\subsection{Experimental setup}
We split the data into three equally sized sets: A training set used to train a prediction model, a calibration set used to calculate non-conformity scores, and a test set used to evaluate the confidence set. The split is done by randomly sampling from the entire data set, without considering the temporal ordering of the data points.\footnote{It is pointed out in \cite{Oust2024} that this is a common approach in testing of AVMs for research purposes, even if it circumvents some issues related to out-of-time performance of the models.} For each individual test instance, we compute a weighted quantile of the calibration scores based on the chosen weighting function and use this to construct the CP set for that particular test point. 

We experiment with four different non-conformity scores and four different configurations of the CP procedure. The non-conformity scores considered are described below. 
\begin{itemize}
    \item \textbf{Standard}: The non-conformity score is the absolute residual, $\lvert Y_i - \hat{f}(X_i)\rvert$.  
    
    \item \textbf{Normalized 1}: The non-conformity score is $\lvert Y_i - \hat{f}(X_i)\rvert/\hat{f}(X_i)$, i.e., the absolute residual normalized by the prediction itself. This is proposed in \cite{Bellotti}, since the absolute residual correlates with the price of the dwelling. 

    \item \textbf{Normalized 2}: The non-conformity score is $\lvert Y_i - \hat{f}(X_i)\rvert/\hat{\rho}(X_i)$, where $\hat{\rho}(X_i)$ is an estimate of the heteroskedasticity in the absolute residuals. The estimate is obtained by fitting a linear model to the unnormalized absolute residuals on a subset of the features. 

    \item \textbf{CQR}: The scores are derived from a quantile regression model, and conformalized in accordance with the description in \autoref{subsec:ncm}. %For the random forests model we use the quantile regression forest implementation proposed in \cite{QRF}, and for the gradient-boosted tree we use the native implementation in LightGBM (\cite{LightGBM}).  
\end{itemize}
Furthermore, we explore four different calibration methods: 
\begin{itemize}
    \item \textbf{CP}: The regular unweighted version of CP. 
    
    \item \textbf{Mondrian CP (MCP)}: The unweighted version of CP, but calibrated per city district. 
    
    \item \textbf{Spatial CP (SCP)}: Weighted CP with weights $w_{i} = \exp (-d_{i,N+1}^2/\eta)$. We choose $\eta$ such that the weight is halved after $800$ meters. 
    
   \item \textbf{Nearest Neighbor CP (NNCP)}: A non-smoothed version of SCP, where $w_{i} = 1$ only in a local neighborhood of size $1$ kilometer around $X_{N+1}$.
\end{itemize}

\subsection{Simulation study}
\label{sec:simulation_study}

Before using the actual sale prices from the Oslo data set, we generate synthetic sale prices from a known error distribution defined over the spatial domain. This allows us to study how each CP method performs in an idealized scenario where the spatial distribution of errors -- and thus non-conformity scores -- can be controlled. 

\subsubsection{Generating responses}
Let $\mathbf{L}_i = (L_1^i, L_2^i) \in [0,1] \times [0,1]$ be the coordinates of observation $i$ in the Oslo data set, normalized to a unit square. We then use the following mechanism to simulate synthetic sale prices:  
\begin{align}
    \Tilde{Y}_i = \hat{f}(X_i) + \varepsilon_i, \quad \varepsilon_i \sim \mathcal{N}(0, \sigma^2(\mathbf{L}_i)), 
    \label{eq:data_generating}
\end{align}
where we use the same function $\hat{f}$ for every observation based on its non-spatial features $X_i$ (which we sample from the real data), and the variance $\sigma_i^2$ depends on the spatial location of the observation. We consider three scenarios for the distribution of the noise: 
\begin{align*}
    \text{Scenario 1}:& \quad\quad \sigma^2(\mathbf{L}_i) = 1, \\ 
    \text{Scenario 2}:& \quad\quad \sigma^2(\mathbf{L}_i) = L_1^i + L_2^i, \\ 
    \text{Scenario 3}:& \quad\quad \sigma^2(\mathbf{L}_i) = \lvert\sin(4\pi L_1^i) + \sin(4\pi L_2^i)\rvert. 
\end{align*}
The first scenario is the simplest scenario where the noise is identically distributed across the spatial domain, regardless of the location. We expect unweighted CP to be sufficient for this scenario, as there should be no spatial trends in the non-conformity scores. In the second scenario, the variance of the error term increases gradually and monotonically in both directions as we move the bottom left corner of the spatial domain (i.e., southwest) towards the top right corner of the spatial domain (i.e., northeast). In the third scenario, the variance of the error term follows a sine function in both directions. Note that the noise is close to zero at some instances in Scenario 3, or even zero at the point masses where both sine functions are zero simultaneously. This scenario is the most challenging, as the variance of the noise term is increasing and decreasing more quickly than in Scenario 2. A visualization of the scenarios are provided in \autoref{appendix:noise}. 

We use a gradient boosted trees model trained on a subset of $5\,000$ samples from the Oslo data set to generate the responses. We only train this model on the non-spatial features of the data set, i.e., leaving out the coordinates and the city district variable. We subsequently discard this part of the data set and use the trained model to construct synthetic sale prices for the rest of the Oslo data set according to \eqref{eq:data_generating}, with the predictions from the gradient boosted trees model serving as  $\hat{f}(X_i)$ based on the features $X_i$ from the Oslo data set. We then proceed with our analysis of the CP methods with the synthetically generated prices $\Tilde{Y}_i$ as the response corresponding to the actual features $X_i$ from the Oslo data set.

\subsubsection{Marginal coverage}
We now present the results of the different combinations of non-conformity scores and calibration methods on the synthetic data set. As a point prediction, we use a gradient boosted trees model with similar parameters to the one used to generate data. Details on the parameters in this model are found in \autoref{appendix:computational_details}. We report the mean of $100$ simulations, where a subset of $N = 5\,000$ samples are used each time. The Root Mean Squared Error (RMSE) of the point prediction model is $1.13$, $1.28$, and $1.01$ in Scenario 1, 2, and 3, respectively. The lower RMSE in Scenario 3 is attributed to the areas with relatively low noise terms. 

%%%%% DON'T DISPLAY THIS, STARTING HERE.... 
\iffalse

\begin{table}[h]
\centering
    \begin{tabular}{@{}lcccc@{}}
    \toprule
        Model & RMSE & MdAE & PER10 ($\%$) & PER20 ($\%$) \\ 
        \midrule
       Scenario 1 & 1.11  & 0.75 & 30.0 & 55.0 \\ 
       Scenario 2 & 1.27 & 0.81 & 28.0 & 51.5 \\  
        Scenario 3 & 0.99 & 0.43 & 49.0 & 72.5  \\ 
        \bottomrule
     \end{tabular}
         \caption{Accuracy of the point predictions in the scenario with synthetically generated sale prices from the gradient boosted tree model. The table displays Root Mean Squared Error (RMSE),  Median Absolute Error (MdAE), and Percentage Error Range (PER$10$ and PER$20$). \textcolor{red}{Consider just writing this in the text, focusing on RMSE.}}
         \label{tab:accuracy_synthetic2}
\end{table}

\fi 
%%%%% ... ENDING HERE

\begin{figure}
    \centering
    \includegraphics[width = 0.99\linewidth]{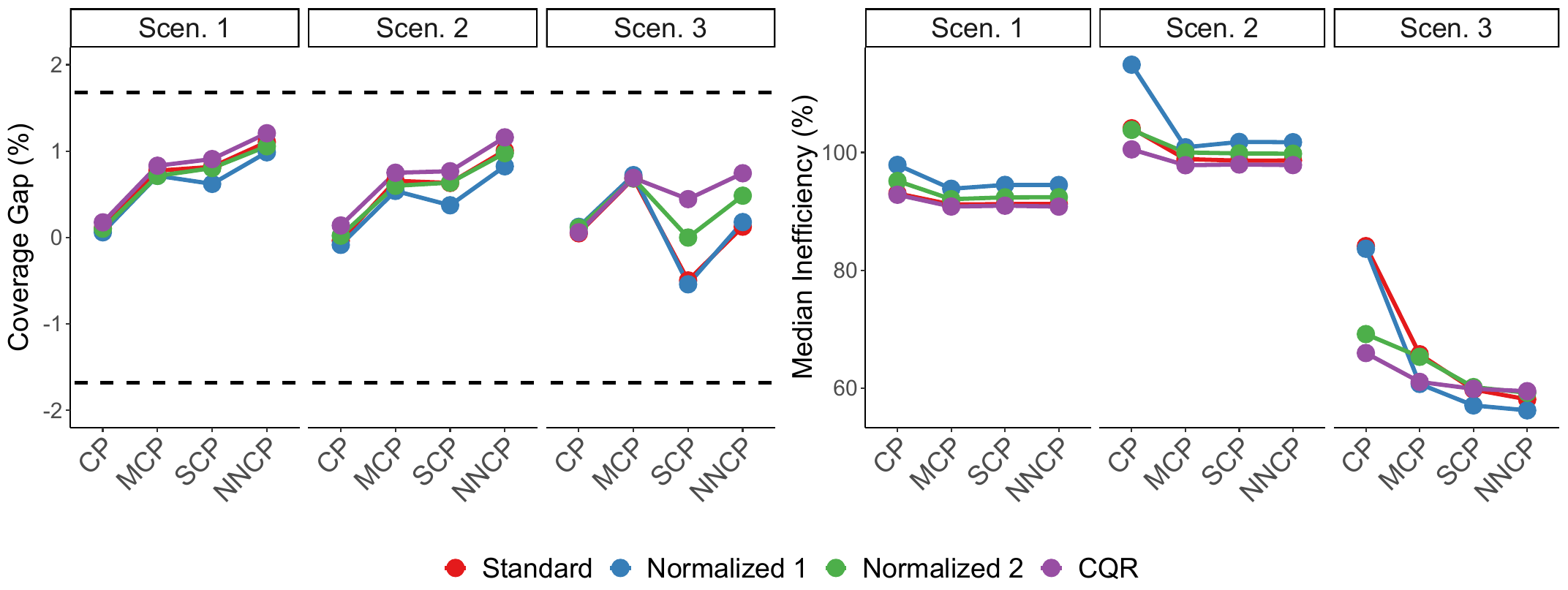}
    \caption{The marginal performance of CP sets at $1-\alpha = 0.9$ on synthetic data set. These results are averaged over $100$ simulations, each time with a new random split into training, calibration, and testing. \textbf{Left}: Empirical coverage gap from the desired $1-\alpha$.  \textbf{Right}: Size of the CP set as a percentage of the (synthetic) sale price.}
    \label{fig:synthetic_marginal_results}
\end{figure}

Moving to the evaluation of the CP sets, 
\autoref{fig:synthetic_marginal_results} summarizes the marginal coverage and inefficiency at confidence level $0.9$ for the combinations of non-conformity scores and weighting methods for each of the three scenarios. As a benchmark for the empirical coverage, we include dotted lines indicating the $5$th and $95$th percentile of the theoretical coverage distribution of a CP set, which follows a known beta-binomial distribution (see \autoref{appendix:beta_binomial} for the exact specification). Interestingly, every combination of non-conformity score and weighting method yields a coverage gap within the expected variability for every noise scenario, highlighting the robustness of the CP framework under different circumstances. For every scenario the CP calibration method yields a coverage gap close to zero for every non-conformity score. For Scenario 1 and 2, the coverage gap increases slightly when we calibrate with MCP, SCP, or NNCP. For Scenario 3, the coverage gap of the SCP and NNCP methods varies more depending on the choice of non-conformity score, although all the methods have an absolute coverage gap of at most $1\%$. 

The right plot of \autoref{fig:synthetic_marginal_results} displays the median inefficiency, with lower values indicating narrower intervals. The inefficiency is rather similar across all methods for the first two scenarios, although there is a slightly higher inefficiency for the CP calibration method. For Scenario 3, it is clear that calibrating with MCP, SCP, or NNCP makes the sets more efficient than CP. This highlights the adaptability of these methods when the magnitude of the non-conformity scores change across the spatial domain.

\subsubsection{Approximate conditional coverage}
To study the approximate conditional coverage in different geographical subsets, we consider the Mean Absolute Coverage Gap (MACG) across geographical subsets as our performance measure, with lower values being preferred. We can either utilize the $15$ city districts that are inherent in the data set or partition the spatial domain into an equidistant grid regardless of city district borders. For the latter, we construct a $20\times 20$ grid over the spatial domain, although, in practice, only approximately half of the grid cells contain observations because of the shape of the studied area. 

\begin{figure}
    \centering
    \includegraphics[width = \linewidth]{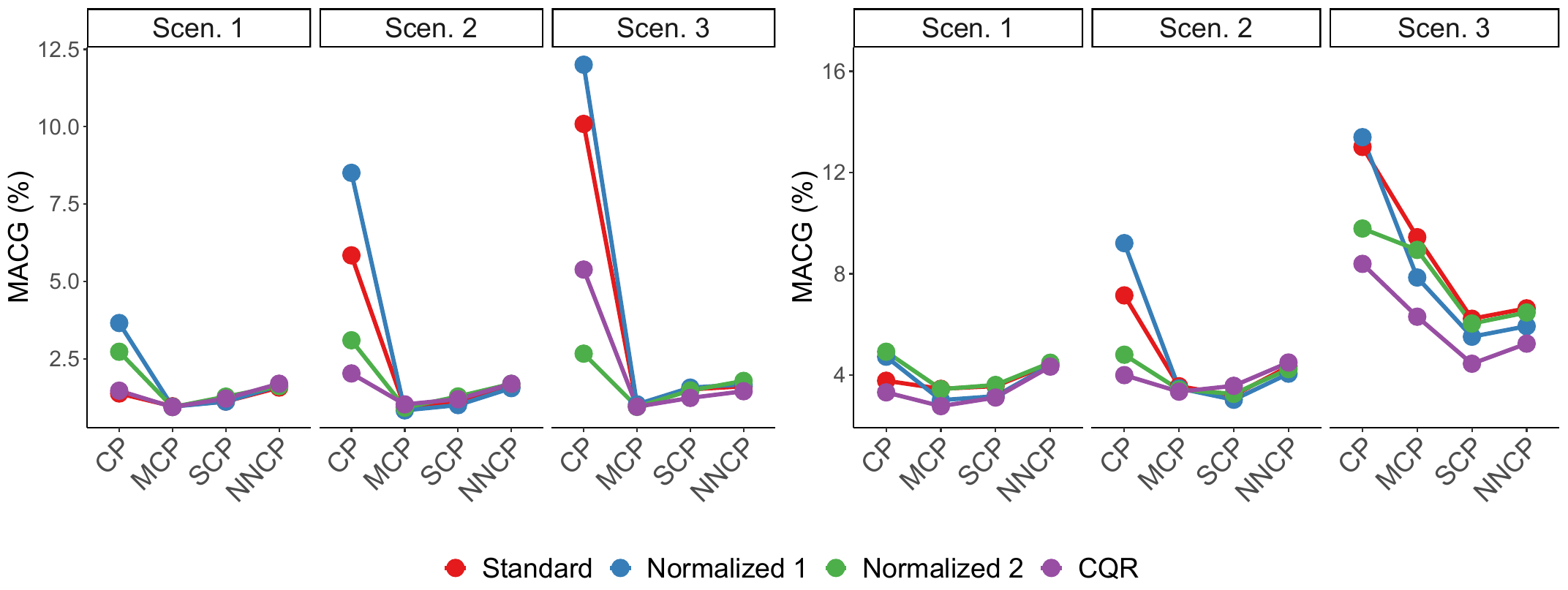}
      \caption{Mean Absolute Coverage Gap across geographical subsets in the data. \textbf{Left:} Evaluating with the city districts in Oslo as subsets. \textbf{Right:} Evaluating with a $20 \times 20$ grid of the spatial domain as subsets. We display the plots with individual $y$ axes, as the two plots are evaluated on a different spatial granularity.}
    \label{fig:macg_synthetic}
\end{figure}

Both of these evaluation methods are depicted in \autoref{fig:macg_synthetic}. Unsurprisingly, the MCP is the best calibration method when using the city districts as our geographical subsets, as it is also calibrated on the same spatial granularity. In this context, there is little to no improvement in using SCP or NNCP instead of MCP. The standard non-conformity score is among the best performers in Scenario 1, highlighting that the simplest approach is sufficient to obtain approximate conditional coverage in a setting where the noise is similar everywhere. 

When we evaluate the methods over a grid in the spatial domain, regardless of the city districts (the right figure in \autoref{fig:macg_synthetic}), all the calibration methods are approximately equal for Scenario 1 and Scenario 2. For Scenario 3, however, there is a clear improvement from CP to SCP and NNCP for every non-conformity score. As one would also expect, the performance of MCP is not as good when we evaluate the performance across a grid instead of the city districts. With this evaluation method, the SCP and NNCP are proving to be the best methods, as they are better suited than MCP to adapt to the quickly changing noise patterns of Scenario 3.

\subsection{Results on the Oslo data set}
\label{subsec:results_real_data}
We now repeat the analysis with actual sale prices instead of synthetically generated ones. We report the mean of $30$ experiments, this time on the complete data set of $N = 26\,362$ observations with the actual sale price. We perform all experiments with both a random forests model and a gradient boosted trees as the point prediction model, however, due to the results being very similar we only report the results with a gradient boosted tree model here. The experiments with a random forests model are reported in \autoref{appendix:randomforests}. \autoref{tab:accuracy_real} displays the predictive accuracy for the actual sale prices. In addition to the RMSE, we also include Median Absolute Error (MdAE) and the PER10 and PER20 measures, which quantify the fraction of the test set within $\pm10\%$ and $\pm 20\%$ of the actual sale price, respectively, as these metrics are often utilized in the AVM industry \citep{SteurerHillPfeifer2021}. The precision of the point predictions is generally higher in the actual data set than in the synthetic case, evident by the lower RMSE of $0.62$. 

\begin{table}[h]
    \centering
    \begin{tabular}{lcccc}
    \toprule
        Model & RMSE & MdAE & PER10 ($\%$) & PER20 ($\%$) \\ 
    \midrule
         Gradient boosted trees & 0.62  & 0.24 & 71.0 & 93.8 \\ 
%        Random forests & 0.64 & 0.24 & 70.4 & 93.6 \\ 
       \bottomrule
         \end{tabular}
    \caption{Accuracy of the point predictions on the real data. The table displays Root Mean Squared Error (RMSE),  Median Absolute Error (MdAE), and Percentage Error Range (PER$10$ and PER$20$).}
    \label{tab:accuracy_real}
\end{table}

\subsubsection{Marginal coverage}

\begin{figure}
    \centering
    \includegraphics[width = \linewidth]{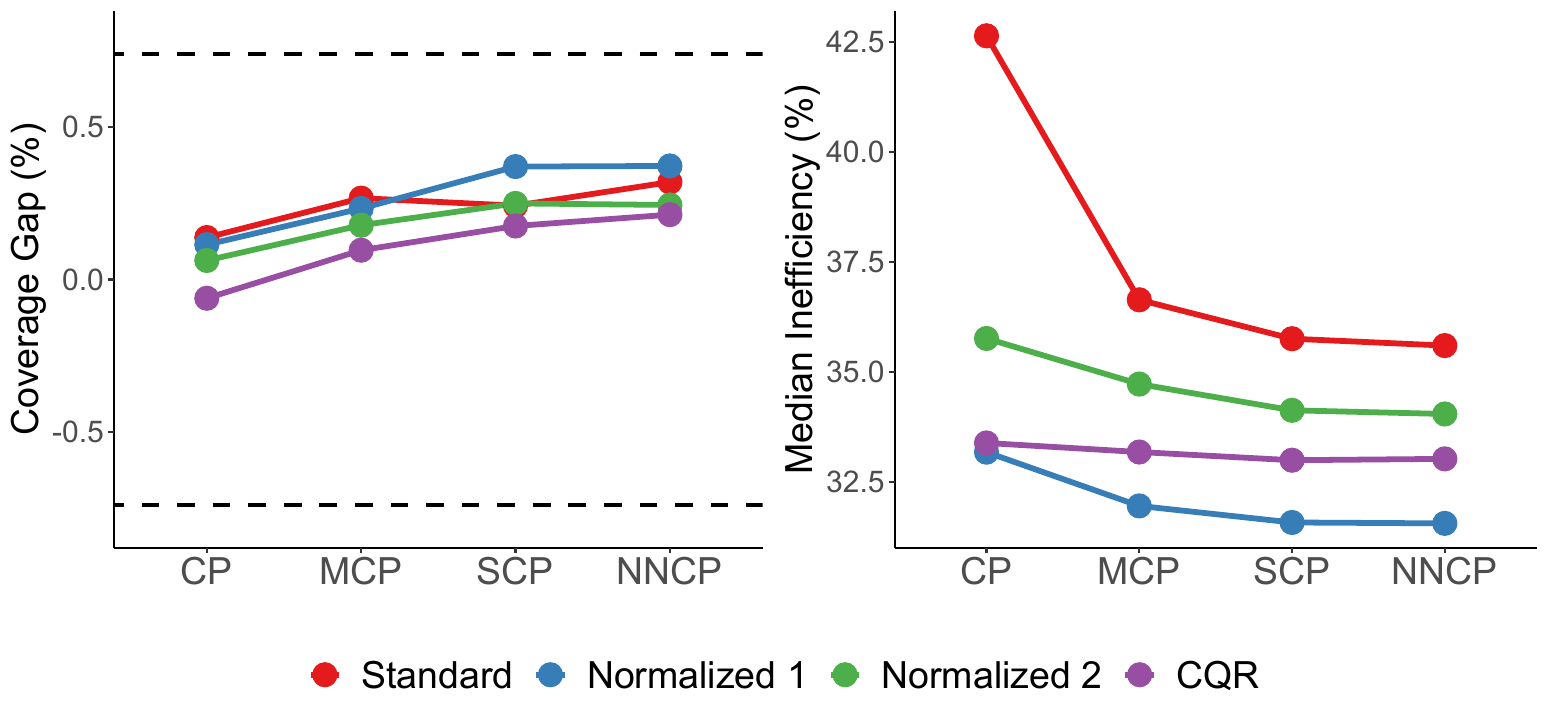}
      \caption{The performance of the CP sets at $1-\alpha = 0.9$ on the real data set. \textbf{Left}: Empirical coverage gap from the desired $1-\alpha$. \textbf{Right}: Size of the CP set as a percentage of the actual sale price.}
    \label{fig:realdata_commonplots}
\end{figure}

The coverage gap and relative efficiency are displayed in \autoref{fig:realdata_commonplots}. The marginal coverage gaps are close to zero and within the expected variability for all methods. Furthermore, the Normalized 1 non-conformity score creates the narrowest intervals for every calibration method with a median inefficiency around $32\%$ of the sale price for every calibration method. This is in line with the findings of \cite{Bellotti}, who mention Normalized 1 as a non-conformity score suited for real estate applications. The CQR constructs consistently efficient sets, with little to no improvement as we change the calibration method. The standard non-conformity score yields the most inefficient intervals, particularly for the CP calibration method, albeit with a clear improvement when using one of the local calibration methods (MCP, SCP, or NNCP).

\subsubsection{Approximate conditional coverage}

\begin{figure}
    \centering
    \includegraphics[width = 0.99\linewidth]{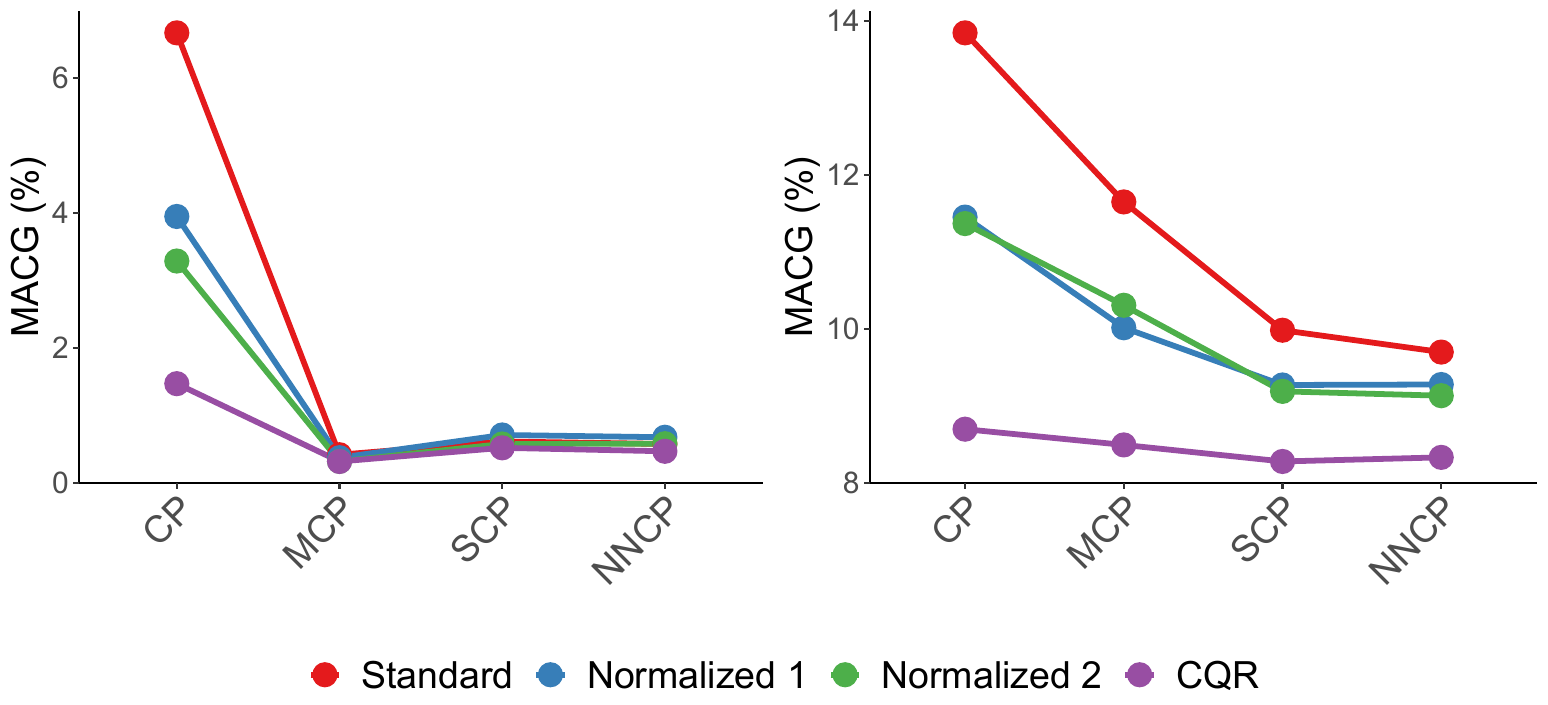}
    \caption{Mean Absolute Coverage Gap (MACG) across geographical subsets. \textbf{Left:} Evaluating with the city districts in Oslo as subsets. \textbf{Right:} Evaluating with a $20 \times 20$ grid of the spatial domain as subsets. We display the plots with different scales on $y$ axis, as the two plots are evaluated on a different spatial granularity.}
    \label{fig:macg}
\end{figure}

In \autoref{fig:macg}, we summarize the approximate conditional coverage through the MACG across different city districts (left plot), and across a $20\times 20$ grid over the spatial domain (right plot). The scale of the $y$ axis differs between the two plots, as the MACG levels are naturally higher when we measure the coverage gap in a smaller spatial domain with fewer observations. 

As in the simulation study, the MCP performs best under the first evaluation method, which is unsurprising given that we calibrate and evaluate on the same spatial granularity. There is no improvement in performance when using SCP or NNCP instead of MCP when evaluating across the city districts. The CQR non-conformity score achieves a much lower MACG than the standard non-conformity score, even with the CP calibration method, indicating that this method is adaptive without spatial calibration methods. 

When evaluating over the grid (the right figure), the performance is further improved by using the SCP or NNCP calibration methods, indicating that these two methods are better suited than MCP to adapt to local patterns in the non-conformity scores. The improvement in performance by calibrating more locally (through SCP or NNCP) is more pronounced with the standard non-conformity score. This confirms that the standard non-conformity score, when calibrated globally, often fails to capture any spatial patterns in the non-conformity scores. On the contrary, the CQR non-conformity score shows almost no improvement when changing calibration methods but performs well even when calibrating globally. The observed robustness of the CQR method supports the conclusions in \cite{BastosPaquette2024}, who also find that the CQR method yields approximate conditional coverage in different bins of the feature space in a real estate context, even if their research did not consider weighted approaches. 

In general, it is hard to clearly distinguish the performance of the SCP and NNCP calibration methods, which give similar results on most of the performance measures studied in this research. This is not surprising since the former can be considered a smoother version of the latter.

Summarizing the results across all the studied performance measures, the use of CQR is generally the most favorable approach, even when calibrating globally. This is a more sophisticated approach than the other non-conformity scores studied in the sense that more of the spatial variability is encoded explicitly in the non-conformity score through the quantile regression estimates. There is less need to adjust the calibration process by using a weighted sample if the non-conformity scores are already incorporating the spatial dynamics present in the real estate data in a satisfying way. On the contrary, the standard non-conformity score with CP calibration generally achieves the worst MACG and the most inefficient sets, even if it is marginally valid. Despite this, the results are improved in terms of inefficiency and approximate conditional coverage when combining the standard non-conformity score with the SCP or NNCP calibration method. As such, these weighting methods often produce useful confidence sets even if the non-conformity score lacks the adaptivity of CQR. 
\section{Discussion}
\label{sec:conclusion}

This research has surveyed various CP methods for the application of house price prediction in the Norwegian housing market. We are motivated by an acknowledgment that the application of off-the-shelf CP methods sometimes leads to uneven coverage across geographical regions, and we investigate both several non-conformity measures and several calibration methods to account for this. All the surveyed methods display an empirical marginal coverage gap within the expected variability. We study approximate conditional coverage by looking at how the empirical coverage gap varies across geographical subsets of the data set. In general, the CQR non-conformity is the most adaptive because it creates intervals with low coverage gaps also in specific geographical subsets. We also find that calibrating based on a local sample of non-conformity scores around a test instance (through the SCP or NNCP calibration method) can create adaptive intervals even if the non-conformity score is more naïve than the CQR score function. 

An improvement in performance through spatially weighted CP can also be seen as an indication of either a misspecified prediction model or a non-conformity score that fails to adequately capture the spatial trends in the data, leading to scores that vary in magnitude across the spatial domain. This instigates questions surrounding the exchangeability of the non-conformity scores and whether the scores should be treated as exchangeable, or rather \textit{locally} exchangeable in a neighborhood around a test instance, as described in \cite{MartinMaoReich}. Weighting methods have been used in the literature to improve approximate conditional coverage, as we have done in this research, but it is also used in \cite{BeyondExchangeability} to relax the exchangeability assumption. The relationship between exchangeability, local exchangeability, and weighting methods in the context of real estate data is a topic that should be further investigated. 

A topic that has not been studied thoroughly in this paper is the temporal drift in house prices, including cyclical and seasonal patterns. While we include temporal fixed effects per month in the prediction model, we do not consider temporally weighted CP \citep{BeyondExchangeability} or adaptive approaches that account for data drifts over time \citep{Gibbs2021}. A natural extension of this work is to investigate the temporal aspects of the housing market data and adapt CP methods accordingly to create approximate conditional coverage along both the spatial and temporal dimensions.

\subsection{Declarations}
\textbf{Funding:} AH was funded by the Research Council of Norway through the Industrial PhD scheme (grant number 322779). JP was supported by the Research Council of Norway through the Centers of Excellence scheme (Integreat, grant number 332645). AH is employed by Eiendomsverdi AS. \\
\textbf{Conflicts of interest/Competing interests:} The housing market data used in the research and leading to this manuscript are owned and have been provided by Eiendomsverdi AS. Eiendomsverdi AS funded AH’s graduate studies and salary during the duration of this research and manuscript preparation, but the research direction and development and the manuscript content are all solely the responsibility of the authors and do not necessarily represent the official views of Eiendomsverdi AS.\\
%\textbf{Ethics approval:} Not Applicable.\\ \textbf{Consent to participate:} Not Applicable. \\
%\textbf{Consent for publication:} Not Applicable. \\
\textbf{Availability of data and material:} The data is owned by Eiendomsverdi AS and is not available to the public. \\
\textbf{Code availability:} The code used to create the results in this paper is available on \url{https://github.com/adhjort/ConformalPrediction}.  \\
\textbf{Authors' contributions:} AH, GHH, JP, and JPW contributed to the conceptualization. AH wrote the main manuscript in collaboration with JPW. AH wrote the code used in the project. AH, GHH, JP, and JPW reviewed the manuscript and contributed to the editing process.

% NATBIB
\bibliographystyle{elsarticle-harv}
\bibliography{sn-bibliography}

\begin{thebibliography}{34}
\expandafter\ifx\csname natexlab\endcsname\relax\def\natexlab#1{#1}\fi
\providecommand{\url}[1]{\texttt{#1}}
\providecommand{\href}[2]{#2}
\providecommand{\path}[1]{#1}
\providecommand{\DOIprefix}{doi:}
\providecommand{\ArXivprefix}{arXiv:}
\providecommand{\URLprefix}{URL: }
\providecommand{\Pubmedprefix}{pmid:}
\providecommand{\doi}[1]{\href{http://dx.doi.org/#1}{\path{#1}}}
\providecommand{\Pubmed}[1]{\href{pmid:#1}{\path{#1}}}
\providecommand{\bibinfo}[2]{#2}
\ifx\xfnm\relax \def\xfnm[#1]{\unskip,\space#1}\fi
%Type = Article
\bibitem[{Bastos and Paquette(2024)}]{BastosPaquette2024}
\bibinfo{author}{Bastos, J.A.}, \bibinfo{author}{Paquette, J.}, \bibinfo{year}{2024}.
\newblock \bibinfo{title}{On the uncertainty of real estate price predictions}.
\newblock \bibinfo{journal}{Journal of Property Research} \bibinfo{volume}{0}, \bibinfo{pages}{1--19}.
%Type = Article
\bibitem[{Bellotti(2017)}]{Bellotti2017}
\bibinfo{author}{Bellotti, A.}, \bibinfo{year}{2017}.
\newblock \bibinfo{title}{Reliable region predictions for automated valuation models}.
\newblock \bibinfo{journal}{Annals of Mathematics and Artificial Intelligence} \bibinfo{volume}{81}, \bibinfo{pages}{71--84}.
%Type = Article
\bibitem[{Bostr{\"o}m et~al.(2017)Bostr{\"o}m, Linusson, L{\"o}fstr{\"o}m and Johansson}]{Bostrom2017}
\bibinfo{author}{Bostr{\"o}m, H.}, \bibinfo{author}{Linusson, H.}, \bibinfo{author}{L{\"o}fstr{\"o}m, T.}, \bibinfo{author}{Johansson, U.}, \bibinfo{year}{2017}.
\newblock \bibinfo{title}{Accelerating difficulty estimation for conformal regression forests}.
\newblock \bibinfo{journal}{Annals of Mathematics and Artificial Intelligence} \bibinfo{volume}{81}, \bibinfo{pages}{125--144}.
%Type = Article
\bibitem[{Breiman(2001)}]{BreimanRF}
\bibinfo{author}{Breiman, L.}, \bibinfo{year}{2001}.
\newblock \bibinfo{title}{Random forests}.
\newblock \bibinfo{journal}{Machine Learning} \bibinfo{volume}{45}, \bibinfo{pages}{5--23}.
%Type = Article
\bibitem[{Chen and Guestrin(2016)}]{XGBoost}
\bibinfo{author}{Chen, T.}, \bibinfo{author}{Guestrin, C.}, \bibinfo{year}{2016}.
\newblock \bibinfo{title}{Xgboost: A scalable tree boosting system}.
\newblock \bibinfo{journal}{Proceedings of the 22nd ACM SIGKDD International Conference on Knowledge Discovery and Data Mining} .
%Type = Article
\bibitem[{Foygel~Barber et~al.(2023)Foygel~Barber, Cand{\`e}s, Ramdas and Tibshirani}]{BeyondExchangeability}
\bibinfo{author}{Foygel~Barber, R.}, \bibinfo{author}{Cand{\`e}s, E.}, \bibinfo{author}{Ramdas, A.}, \bibinfo{author}{Tibshirani, R.J.}, \bibinfo{year}{2023}.
\newblock \bibinfo{title}{{Conformal prediction beyond exchangeability}}.
\newblock \bibinfo{journal}{The Annals of Statistics} \bibinfo{volume}{51}, \bibinfo{pages}{816 -- 845}.
%Type = Article
\bibitem[{Foygel~Barber et~al.(2020)Foygel~Barber, Candès, Ramdas and Tibshirani}]{LimitsOfDistributionFreeCoverage}
\bibinfo{author}{Foygel~Barber, R.}, \bibinfo{author}{Candès, E.J.}, \bibinfo{author}{Ramdas, A.}, \bibinfo{author}{Tibshirani, R.J.}, \bibinfo{year}{2020}.
\newblock \bibinfo{title}{The limits of distribution-free conditional predictive inference}.
\newblock \bibinfo{journal}{Information and Inference: A Journal of the IMA} \bibinfo{volume}{10}, \bibinfo{pages}{455--482}.
%Type = Inproceedings
\bibitem[{Gibbs and Cand{\`e}s(2021)}]{Gibbs2021}
\bibinfo{author}{Gibbs, I.}, \bibinfo{author}{Cand{\`e}s, E.J.}, \bibinfo{year}{2021}.
\newblock \bibinfo{title}{Adaptive conformal inference under distribution shift}, in: \bibinfo{booktitle}{Neural Information Processing Systems}, pp. \bibinfo{pages}{1660--1672}.
%Type = Article
\bibitem[{Guan(2022)}]{Guan}
\bibinfo{author}{Guan, L.}, \bibinfo{year}{2022}.
\newblock \bibinfo{title}{{Localized conformal prediction: a generalized inference framework for conformal prediction}}.
\newblock \bibinfo{journal}{Biometrika} \bibinfo{volume}{110}, \bibinfo{pages}{33--50}.
%Type = Article
\bibitem[{Hao et~al.(2023)Hao, Lin, Shen and Su}]{Hao2023}
\bibinfo{author}{Hao, M.}, \bibinfo{author}{Lin, Y.}, \bibinfo{author}{Shen, G.}, \bibinfo{author}{Su, W.}, \bibinfo{year}{2023}.
\newblock \bibinfo{title}{Nonparametric inference on smoothed quantile regression process}.
\newblock \bibinfo{journal}{Computational Statistics \& Data Analysis} \bibinfo{volume}{179}, \bibinfo{pages}{107645}.
%Type = Article
\bibitem[{Hjort et~al.(2022)Hjort, Pensar, Scheel and Sommervoll}]{Hjort2022}
\bibinfo{author}{Hjort, A.}, \bibinfo{author}{Pensar, J.}, \bibinfo{author}{Scheel, I.}, \bibinfo{author}{Sommervoll, D.E.}, \bibinfo{year}{2022}.
\newblock \bibinfo{title}{House price prediction with gradient boosted trees under different loss functions}.
\newblock \bibinfo{journal}{Journal of Property Research} \bibinfo{volume}{39}, \bibinfo{pages}{338--364}.
%Type = Inproceedings
\bibitem[{Hjort et~al.(2024)Hjort, Williams and Pensar}]{Hjort2024}
\bibinfo{author}{Hjort, A.}, \bibinfo{author}{Williams, J.P.}, \bibinfo{author}{Pensar, J.}, \bibinfo{year}{2024}.
\newblock \bibinfo{title}{Clustered conformal prediction for the housing market}, in: \bibinfo{editor}{Vantini, S.}, \bibinfo{editor}{Fontana, M.}, \bibinfo{editor}{Solari, A.}, \bibinfo{editor}{Boström, H.}, \bibinfo{editor}{Carlsson, L.} (Eds.), \bibinfo{booktitle}{Proceedings of the Thirteenth Symposium on Conformal and Probabilistic Prediction with Applications}, \bibinfo{publisher}{PMLR}. pp. \bibinfo{pages}{366--386}.
%Type = Techreport
\bibitem[{van Hoenselaar et~al.(2021)van Hoenselaar, Cournède, Pace and Ziemann}]{OECD_MortgageFinance}
\bibinfo{author}{van Hoenselaar, F.}, \bibinfo{author}{Cournède, B.}, \bibinfo{author}{Pace, F.D.}, \bibinfo{author}{Ziemann, V.}, \bibinfo{year}{2021}.
\newblock \bibinfo{title}{Mortgage finance across OECD countries}.
\newblock \bibinfo{type}{Technical Report} \bibinfo{number}{1693}. Organization for Economic Cooperation and Development.
%Type = Article
\bibitem[{Johansson et~al.(2022)Johansson, Sönströd, Löfström and Boström}]{Johansson2022}
\bibinfo{author}{Johansson, U.}, \bibinfo{author}{Sönströd, C.}, \bibinfo{author}{Löfström, T.}, \bibinfo{author}{Boström, H.}, \bibinfo{year}{2022}.
\newblock \bibinfo{title}{Rule extraction with guarantees from regression models}.
\newblock \bibinfo{journal}{Pattern Recognition} \bibinfo{volume}{126}, \bibinfo{pages}{108554}.
%Type = Inproceedings
\bibitem[{Ke et~al.(2017)Ke, Meng, Finley, Wang, Chen, Ma, Ye and Liu}]{LightGBM}
\bibinfo{author}{Ke, G.}, \bibinfo{author}{Meng, Q.}, \bibinfo{author}{Finley, T.}, \bibinfo{author}{Wang, T.}, \bibinfo{author}{Chen, W.}, \bibinfo{author}{Ma, W.}, \bibinfo{author}{Ye, Q.}, \bibinfo{author}{Liu, T.Y.}, \bibinfo{year}{2017}.
\newblock \bibinfo{title}{Lightgbm: A highly efficient gradient boosting decision tree}, in: \bibinfo{booktitle}{Advances in Neural Information Processing Systems}, p. \bibinfo{pages}{3149–3157}.
%Type = Article
\bibitem[{Krause et~al.(2020)Krause, Martin and Fix}]{Krause2020}
\bibinfo{author}{Krause, A.}, \bibinfo{author}{Martin, A.}, \bibinfo{author}{Fix, M.}, \bibinfo{year}{2020}.
\newblock \bibinfo{title}{Uncertainty in automated valuation models: Error-based versus model-based approaches}.
\newblock \bibinfo{journal}{Journal of Property Research} \bibinfo{volume}{37}, \bibinfo{pages}{308--339}.
%Type = Article
\bibitem[{Lei et~al.(2018)Lei, Rinaldo, Tibshirani, G'Sell and Wasserman}]{lei2017distributionfree}
\bibinfo{author}{Lei, J.}, \bibinfo{author}{Rinaldo, A.}, \bibinfo{author}{Tibshirani, R.J.}, \bibinfo{author}{G'Sell, M.}, \bibinfo{author}{Wasserman, L.}, \bibinfo{year}{2018}.
\newblock \bibinfo{title}{Distribution-free predictive inference for regression}.
\newblock \bibinfo{journal}{Journal of the American Statistical Association} \bibinfo{volume}{113}, \bibinfo{pages}{1094--1111}.
%Type = Article
\bibitem[{Lei and Wasserman(2014)}]{LeiWasserman2014}
\bibinfo{author}{Lei, J.}, \bibinfo{author}{Wasserman, L.}, \bibinfo{year}{2014}.
\newblock \bibinfo{title}{Distribution-free prediction bands for non-parametric regression}.
\newblock \bibinfo{journal}{Journal of the Royal Statistical Society: Series B (Statistical Methodology)} \bibinfo{volume}{76}.
%Type = Article
\bibitem[{Lim and Bellotti(2021)}]{Bellotti}
\bibinfo{author}{Lim, Z.}, \bibinfo{author}{Bellotti, A.}, \bibinfo{year}{2021}.
\newblock \bibinfo{title}{Normalized nonconformity measures for automated valuation models}.
\newblock \bibinfo{journal}{Expert Systems with Applications} \bibinfo{volume}{180}, \bibinfo{pages}{115--165}.
%Type = Article
\bibitem[{Ma et~al.(2020)Ma, Liu, Cao, Song, Zhang and Zeng}]{DeepForest2020}
\bibinfo{author}{Ma, C.}, \bibinfo{author}{Liu, Z.}, \bibinfo{author}{Cao, Z.}, \bibinfo{author}{Song, W.}, \bibinfo{author}{Zhang, J.}, \bibinfo{author}{Zeng, W.}, \bibinfo{year}{2020}.
\newblock \bibinfo{title}{Cost-sensitive deep forest for price prediction}.
\newblock \bibinfo{journal}{Pattern Recognition} \bibinfo{volume}{107}, \bibinfo{pages}{107499}.
%Type = Article
\bibitem[{Mao et~al.(2023)Mao, Martin and Reich}]{MartinMaoReich}
\bibinfo{author}{Mao, H.}, \bibinfo{author}{Martin, R.}, \bibinfo{author}{Reich, B.J.}, \bibinfo{year}{2023}.
\newblock \bibinfo{title}{Valid model-free spatial prediction}.
\newblock \bibinfo{journal}{Journal of the American Statistical Association} \bibinfo{volume}{119}, \bibinfo{pages}{904--914}.
%Type = Inproceedings
\bibitem[{Marques et~al.(2021)Marques, Batista, Castro and Bhattacharjee}]{Marques2021}
\bibinfo{author}{Marques, J.a.L.}, \bibinfo{author}{Batista, P.}, \bibinfo{author}{Castro, E.A.}, \bibinfo{author}{Bhattacharjee, A.}, \bibinfo{year}{2021}.
\newblock \bibinfo{title}{Spatial automated valuation model (savm) – from the notion of space to the design of an evaluation tool}, in: \bibinfo{booktitle}{Computational Science and Its Applications – ICCSA 2021: 21st International Conference, Cagliari, Italy, September 13–16, 2021, Proceedings, Part IV}, \bibinfo{publisher}{Springer-Verlag}, \bibinfo{address}{Berlin, Heidelberg}. p. \bibinfo{pages}{75–90}.
%Type = Article
\bibitem[{Matiz and Barner(2019)}]{Matiz2019}
\bibinfo{author}{Matiz, S.}, \bibinfo{author}{Barner, K.E.}, \bibinfo{year}{2019}.
\newblock \bibinfo{title}{Inductive conformal predictor for convolutional neural networks: Applications to active learning for image classification}.
\newblock \bibinfo{journal}{Pattern Recognition} \bibinfo{volume}{90}, \bibinfo{pages}{172--182}.
%Type = Article
\bibitem[{Messoudi et~al.(2021)Messoudi, Destercke and Rousseau}]{Messoudi2021}
\bibinfo{author}{Messoudi, S.}, \bibinfo{author}{Destercke, S.}, \bibinfo{author}{Rousseau, S.}, \bibinfo{year}{2021}.
\newblock \bibinfo{title}{Copula-based conformal prediction for multi-target regression}.
\newblock \bibinfo{journal}{Pattern Recognition} \bibinfo{volume}{120}, \bibinfo{pages}{108101}.
%Type = Article
\bibitem[{Nie et~al.(2020)Nie, Guo, Chang, Han, Huang, Hu and Zhang}]{Shallow2Deep2020}
\bibinfo{author}{Nie, Y.}, \bibinfo{author}{Guo, S.}, \bibinfo{author}{Chang, J.}, \bibinfo{author}{Han, X.}, \bibinfo{author}{Huang, J.}, \bibinfo{author}{Hu, S.M.}, \bibinfo{author}{Zhang, J.J.}, \bibinfo{year}{2020}.
\newblock \bibinfo{title}{Shallow2deep: Indoor scene modeling by single image understanding}.
\newblock \bibinfo{journal}{Pattern Recognition} \bibinfo{volume}{103}, \bibinfo{pages}{107271}.
%Type = Article
\bibitem[{Oust and Steininger(2024)}]{Oust2024}
\bibinfo{author}{Oust, A.}, \bibinfo{author}{Steininger, B.}, \bibinfo{year}{2024}.
\newblock \bibinfo{title}{3. automated valuation model in real estate}.
\newblock \bibinfo{journal}{Fastighetsv{\"a}rdering 2024: En antologi} .
%Type = Inproceedings
\bibitem[{Papadopoulos et~al.(2002)Papadopoulos, Proedrou, Vovk and Gammerman}]{Papadopoulos2002}
\bibinfo{author}{Papadopoulos, H.}, \bibinfo{author}{Proedrou, K.}, \bibinfo{author}{Vovk, V.}, \bibinfo{author}{Gammerman, A.}, \bibinfo{year}{2002}.
\newblock \bibinfo{title}{Inductive confidence machines for regression}, in: \bibinfo{booktitle}{Machine Learning: ECML 2002}, pp. \bibinfo{pages}{345--356}.
%Type = Inproceedings
\bibitem[{Romano et~al.(2019)Romano, Patterson and Candes}]{romano2019conformalized}
\bibinfo{author}{Romano, Y.}, \bibinfo{author}{Patterson, E.}, \bibinfo{author}{Candes, E.}, \bibinfo{year}{2019}.
\newblock \bibinfo{title}{Conformalized quantile regression}, in: \bibinfo{booktitle}{Advances in Neural Information Processing Systems}.
%Type = Article
\bibitem[{Shafer and Vovk(2008)}]{ShaferVovk2007}
\bibinfo{author}{Shafer, G.}, \bibinfo{author}{Vovk, V.}, \bibinfo{year}{2008}.
\newblock \bibinfo{title}{A tutorial on conformal prediction}.
\newblock \bibinfo{journal}{Journal of Machine Learning Research} \bibinfo{volume}{9}, \bibinfo{pages}{371--421}.
%Type = Article
\bibitem[{Steurer et~al.(2021)Steurer, Hill and Pfeifer}]{SteurerHillPfeifer2021}
\bibinfo{author}{Steurer, M.}, \bibinfo{author}{Hill, R.J.}, \bibinfo{author}{Pfeifer, N.}, \bibinfo{year}{2021}.
\newblock \bibinfo{title}{Metrics for evaluating the performance of machine learning based automated valuation models}.
\newblock \bibinfo{journal}{Journal of Property Research} \bibinfo{volume}{38}, \bibinfo{pages}{99--129}.
%Type = Inproceedings
\bibitem[{Tibshirani et~al.(2019)Tibshirani, Foygel~Barber, Candes and Ramdas}]{CovariateShift}
\bibinfo{author}{Tibshirani, R.J.}, \bibinfo{author}{Foygel~Barber, R.}, \bibinfo{author}{Candes, E.}, \bibinfo{author}{Ramdas, A.}, \bibinfo{year}{2019}.
\newblock \bibinfo{title}{Conformal prediction under covariate shift}, in: \bibinfo{booktitle}{Advances in Neural Information Processing Systems}.
%Type = Article
\bibitem[{Toccaceli(2022)}]{Toccaceli2022}
\bibinfo{author}{Toccaceli, P.}, \bibinfo{year}{2022}.
\newblock \bibinfo{title}{Introduction to conformal predictors}.
\newblock \bibinfo{journal}{Pattern Recognition} \bibinfo{volume}{124}, \bibinfo{pages}{108507}.
%Type = Inproceedings
\bibitem[{Vovk(2012)}]{Vovk2012}
\bibinfo{author}{Vovk, V.}, \bibinfo{year}{2012}.
\newblock \bibinfo{title}{Conditional validity of inductive conformal predictors}, in: \bibinfo{booktitle}{Proceedings of the Asian Conference on Machine Learning}, pp. \bibinfo{pages}{475--490}.
%Type = Book
\bibitem[{Vovk et~al.(2005)Vovk, Gammerman and Shafer}]{ALRW}
\bibinfo{author}{Vovk, V.}, \bibinfo{author}{Gammerman, A.}, \bibinfo{author}{Shafer, G.}, \bibinfo{year}{2005}.
\newblock \bibinfo{title}{Algorithmic Learning in a Random World}.
\newblock \bibinfo{publisher}{Springer-Verlag}, \bibinfo{address}{Berlin, Heidelberg}.

\end{thebibliography}

% BIBLATEX
%\addbibresource{sn-bibliography.bib}
%\printbibliography

%\appendix
\newpage

\begin{appendices}

\section{Computational details}
\label{appendix:computational_details}

%This process is repeated $10$ times for our simulation study. For each of the $10$ synthetic data sets in our simulation study, we construct non-conformity scores using predictions from fitting a gradient boosted trees model (via the XGBoost package presented in \citet{XGBoost}) 

All experiments are conducted in \texttt{R}. For the gradient-boosted trees we use the \texttt{LightGBM} implementation (\cite{LightGBM}) with $M = 1\,000$ trees, a maximum depth of $4$ in each tree, and a learning rate of $0.03$. For the random forests model we use the \texttt{ranger} implementation with $M = 500$ trees. For the non-conformity score Normalized 2, we fit a linear model $\hat{\sigma}(X)$ to the in-sample absolute residuals from the training data, using only the size of the dwelling and a fixed effect for each month. This model is kept deliberately simple to avoid overfitting, given that we are in effect fitting the residuals from a previous model. We use the native \texttt{lm} package for this purpose.

%The spatial regression incorporates a covariance matrix between the observation which is similar to the true data-generating mechanism, i.e., $Y \sim \mathcal{N}_{N}(X\beta, \Sigma)$ with covariance matrix  
$
%\Sigma_{ij} = \sigma_{\varepsilon}^2 + \sigma^2\cdot e^{- \rho \lvert L_i - L_j\rvert},
$
%for $i,j \in \{1,\dots,N\}$, where $L_i$ and $L_j$ are the spatial location of observation $i$ and $j$, and  $\sigma_{\varepsilon}^2$, $\sigma^2$, and $\rho$ are parameters to be estimated based on the training data. 

%The training of the spatial regression consists of finding $\hat{\beta}$ estimates from the training data while taking the spatial structure into account. The estimates are found using a profiled approach, as presented in \cite{schabenberger2004statistical} (Section 5.2.2). The latter method is an Oracle prediction method in the sense that it is based on the true data-generating mechanism, however, it does not have access to the parameter values used to generate the data but estimates them as part of the training process.

The calculation of the weighted quantile uses the \texttt{weighted.quantile} function from the \texttt{modi} package. For the Nearest Neighbor CP, we increase the search radius by $500$ meters if no neighbours are found, however this rarely happens.  

\section{Noise scenarios from simulation study}
\label{appendix:noise}
\begin{figure}[H]
    \centering
    \includegraphics[width=\linewidth]{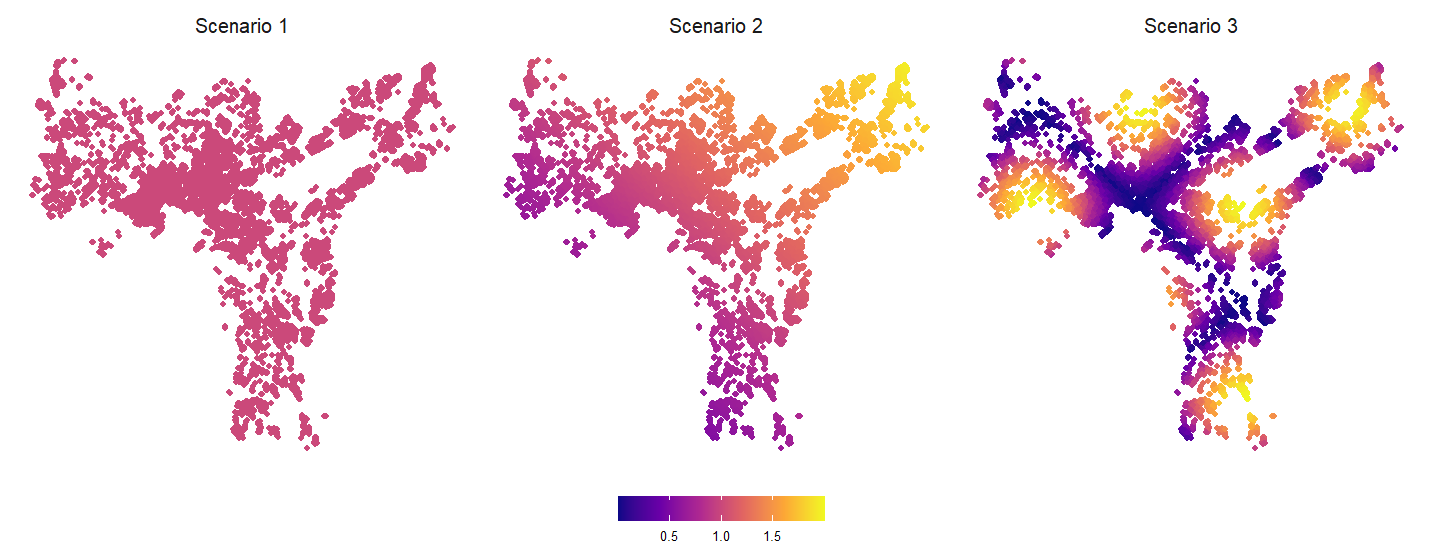}
    \caption{Location-specific variance of the three noise scenarios studied in the paper. Each dot represents an observation in the Oslo data set, such that the position of the dot represents the actual location of the dwelling, and the color of the dot represents the variance at that specific location.}
    \label{fig:noise_scenarios}
\end{figure}

In \autoref{sec:simulation_study}, we construct the following three different  noise scenarios based on the normalized coordinates $\mathbf{L}_i \in [0,1]\times[0,1]$ of observation $i$: 

\begin{align*}
    \text{Scenario 1}:& \quad\quad \sigma^2(\mathbf{L}_i) = 1, \\ 
    \text{Scenario 2}:& \quad\quad \sigma^2(\mathbf{L}_i) = L_1^i + L_2^i, \\ 
    \text{Scenario 3}:& \quad\quad \sigma^2(\mathbf{L}_i) = \lvert\sin(4\pi L_1^i) + \sin(4\pi L_2^i)\rvert. 
\end{align*}
In \autoref{fig:noise_scenarios}, a visualization of these noise scenarios across the data points in the Oslo data set is shown. 

\section{Distribution of coverage}
\label{appendix:beta_binomial}
It is shown in \cite{Vovk2012} that the coverage in an inductive CP setting follows a beta-binomial distribution,
\begin{align*}
    \text{Coverage}(Z_{N+1}, ... ,Z_{N+\Ntest};\alpha) &\sim \frac{1}{\Ntest} \text{Binom}(\Ntest, \mu) \\ 
    \mu &\sim \text{Beta}(n + 1 - l, l), 
\end{align*}
where $l = \floor{\alpha(n+1)}$. Given a number of test points $\Ntest$, a number of calibration points $n$, and a confidence level $\alpha$, we can thus calculate the expected variability in empirical coverage. 

\section{Results with a random forests model}
\label{appendix:randomforests}

\begin{table}
    \centering
    \begin{tabular}{lcccc}
    \toprule
    Model & RMSE & MdAE & PER10 ($\%$) & PER20 ($\%$) \\ 
    \midrule
    Gradient boosted trees & 0.62  & 0.24 & 71.0 & 93.8 \\ 
    Random forests & 0.64 & 0.24 & 70.4 & 93.6 \\ 
   \bottomrule
    \end{tabular}
    \caption{Accuracy of the point predictions on the real data with a random forests model in addition the gradient boosted trees model depicted in the main text. The table displays Root Mean Squared Error (RMSE),  Median Absolute Error (MdAE), and Percentage Error Range (PER$10$ and PER$20$).}
    \label{table:accuracy_rf}
\end{table}

\begin{figure}
    \centering
    \includegraphics[width = 0.9\linewidth]{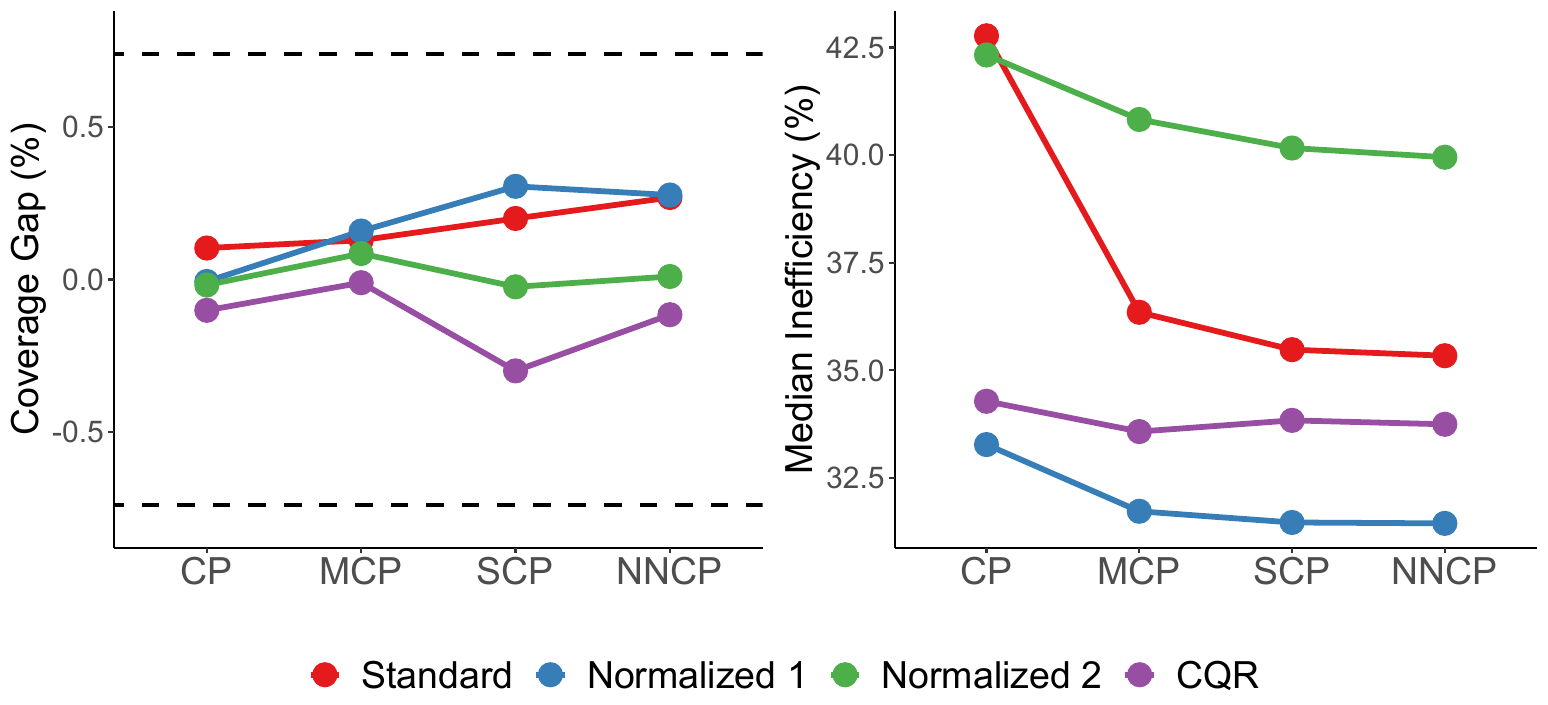}
      \caption{The performance of the CP sets at $1-\alpha = 0.9$ on the real data set. The results are averaged over $30$ simulations. \textbf{Left}: Empirical coverage gap from the desired $1-\alpha$. \textbf{Right}: Size of the CP set as a percentage of the actual sale price.}
    \label{fig:marginalcoverage_rf}
\end{figure}

\begin{figure}
    \centering
    \includegraphics[width = 0.99\linewidth]{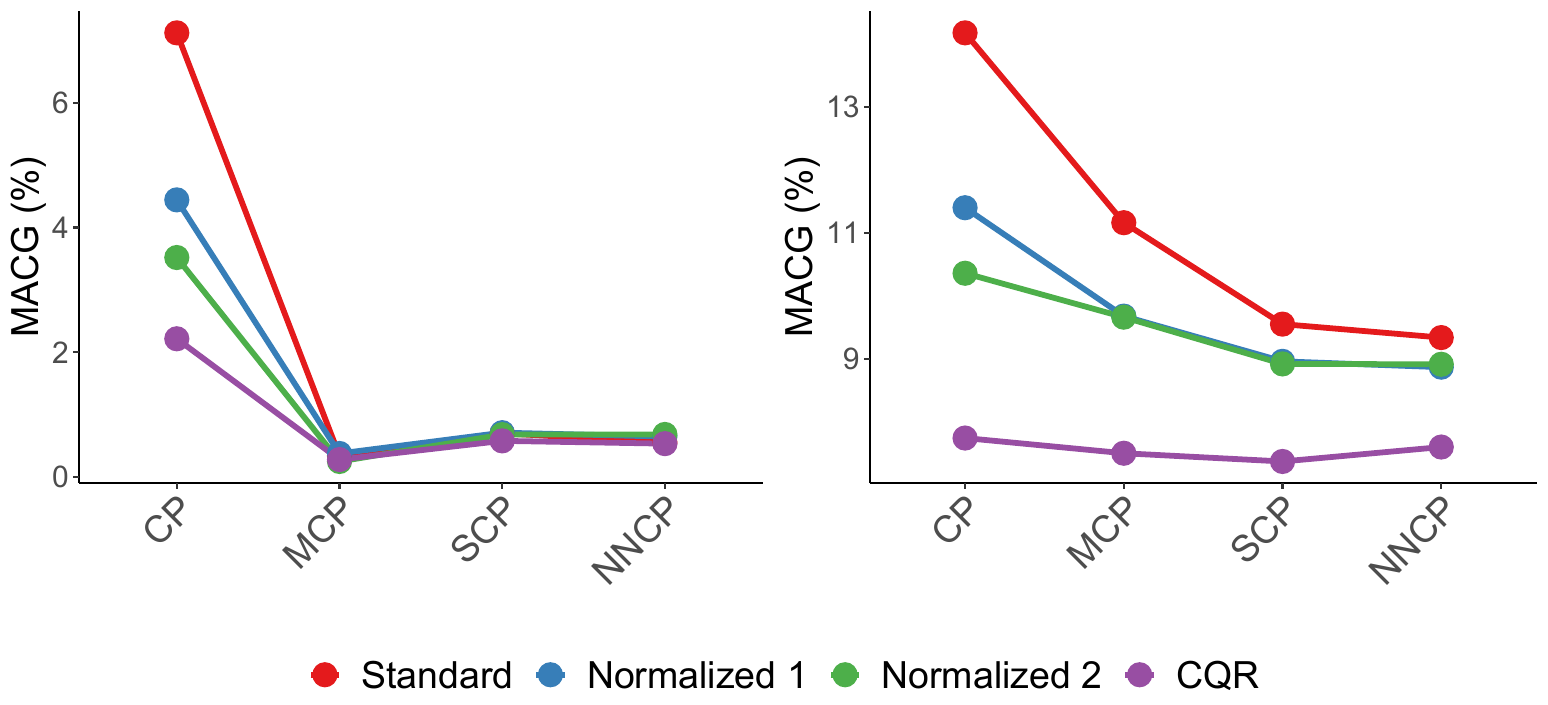}
    \caption{Mean Absolute Coverage Gap across geographical subsets.  \textbf{Left:} Evaluating with the city districts of Oslo as our subsets. \textbf{Right:} Evaluating with a $20 \times 20$ grid of the spatial domain as our subsets. We display the plots with individual $y$ axis to better compare the methods within each scenario. }
    \label{fig:macg_rf}
\end{figure}

We now display the results of the analysis with the actual sale prices, but this time with a random forest model instead of a gradient boosted tree model. In \autoref{table:accuracy_rf}, we display the accuracy of the points prediction models. In \autoref{fig:marginalcoverage_rf}, we display the marginal coverage gap and median inefficiency. In \autoref{fig:macg_rf}, we display the Mean Absolute Coverage Gap (MACG).

\end{appendices}

\end{document}